\documentclass[11pt,letterpaper]{article}

\usepackage{csquotes}

\usepackage[title]{appendix}

\usepackage{caption}
\usepackage{subcaption}
\usepackage{fullpage}
\usepackage{times}
\usepackage{array}
\usepackage{ragged2e}
\usepackage{multirow}
\usepackage{fancyhdr, amssymb, mathtools}
\usepackage[ruled,vlined]{algorithm2e}
\usepackage[dvipsnames,table]{xcolor}
\usepackage[margin = 1 in]{geometry}
\usepackage{graphicx}
\usepackage{outlines}
\usepackage{amsmath}

\usepackage{graphicx}
\usepackage{dirtytalk}

\usepackage{multirow}
\usepackage{arydshln}

\usepackage[english]{babel} 
\usepackage{blindtext}
\usepackage{amsfonts}
\usepackage{amssymb}
\usepackage{bbm}
\usepackage{mathtools}

\usepackage{hhline}
\usepackage{makecell}

\usepackage{xltabular}
\usepackage{arydshln} 
\usepackage{wrapfig}
\usepackage{enumitem}
\usepackage{floatrow}
\usepackage{xfrac}
\usepackage[font={footnotesize}, labelfont={bf}, labelsep=space, justification=justified, skip=0pt]{caption}
\usepackage[hang,flushmargin]{footmisc} 
\usepackage[colorlinks=true, allcolors=blue]{hyperref}

\usepackage{booktabs}

\definecolor{headcolor}{RGB}{255, 255, 255}
\definecolor{columncolor}{RGB}{255, 255, 255}
\definecolor{modelcolor}{RGB}{255, 255, 255}
\definecolor{optioncolor}{RGB}{255, 255, 255}

\definecolor{codegreen}{rgb}{0,0.6,0}
\definecolor{codegray}{rgb}{0.5,0.5,0.5}
\definecolor{codepurple}{rgb}{0.58,0,0.82}
\definecolor{backcolour}{rgb}{0.95,0.95,0.92}

\usepackage{listings}
\lstdefinestyle{mystyle}{
    backgroundcolor=\color{backcolour},   
    commentstyle=\color{codegreen},
    keywordstyle=\color{magenta},
    numberstyle=\tiny\color{codegray},
    stringstyle=\color{codepurple},
    basicstyle=\ttfamily\footnotesize,
    breakatwhitespace=false,         
    breaklines=true,                 
    captionpos=b,                    
    keepspaces=true,                 
    numbers=left,                    
    numbersep=5pt,                  
    showspaces=false,                
    showstringspaces=false,
    showtabs=false,                  
    tabsize=2
}
\usepackage{cleveref} 
\usepackage{cite} 

\AtBeginEnvironment{appendices}{\crefalias{section}{appendix}}

\newcommand\parens[1]{\mathopen{}\left(#1\right)\mathclose{}}
\newcommand\braces[1]{\mathopen{}\left\{#1\right\}\mathclose{}}

\newcommand{\cmt}[1]{} 

\DeclareMathOperator*{\E}{\mathbb{E}}

\newcommand{\alloy}{17-4 PH SS}

\newcommand{\beginsupplement}{%
        \setcounter{table}{0}
        \renewcommand{\thetable}{S\arabic{table}}%
        \setcounter{figure}{0}
        \renewcommand{\thefigure}{S\arabic{figure}}%
     }

\newcommand{\xsspace}{\mathbb{X}}
\newcommand{\rsspace}{\mathbb{R}}

\newcommand{\betab}{\boldsymbol{\beta}}

\newcommand{\thetab}{\boldsymbol{\theta}}

\newcommand{\omegab}{\boldsymbol{\omega}}
\newcommand{\gp}{\texttt{GP+}}
\newcommand{\yb}{\boldsymbol{y}}
\newcommand{\ub}{\boldsymbol{u}}
\newcommand{\tb}{\boldsymbol{t}}

\newcommand{\xb}{\boldsymbol{x}}
\newcommand{\Xb}{\boldsymbol{X}}
\newcommand{\Ib}{\boldsymbol{I}}

\newcommand{\hb}{\boldsymbol{h}}
\newcommand{\mb}{\boldsymbol{m}}

\newcommand{\Cb}{\boldsymbol{C}}

\newcommand{\pib}{\boldsymbol{\pi}}

\newcommand{\EP}{$y_{EP}$} 
\newcommand{\hh}{$\hat{y}_H$} 
\newcommand{\hp}{$\hat{y}_{EP}$}  
\newcommand{\hy}{$\hat{y}_{\sigma_Y}$}  
\newcommand{\hd}{$\hat{y}_{\varepsilon_f}$}    
\newcommand{\yh}{$y_H$}
\newcommand{\yp}{$y_P$}
\newcommand{\yy}{$y_{\sigma_Y}$}
\newcommand{\yd}{$y_{\varepsilon_f}$}
\newcommand{\YS}{$\sigma_Y$}
\newcommand{\UTS}{$\sigma_U$}
\newcommand{\D}{$\varepsilon_f$}

\usepackage{titling}
\thanksheadextra{}{}
\thanksheadextra{}{} 
\usepackage{lipsum} 

{\title{\fontsize{14}{14}\selectfont 
Unveiling Processing--Property Relationships in Laser Powder Bed Fusion: The Synergy of Machine Learning and High-throughput Experiments
}}

\date{\vspace{-5ex}}
\usepackage{authblk}
\author[1]{Mahsa Amiri\thanks{\noindent Equal contribution}}
\author[*2]{Zahra Zanjani Foumani}
\author[1,2,3]{Penghui Cao}
\author[1,2,3]{Lorenzo Valdevit\thanks{\noindent Corresponding Authors: valdevit@uci.edu  and raminb@uci.edu}}
\author[$\dagger$2, 4]{Ramin Bostanabad}
\affil[1]{Materials and Manufacturing Technology Program, University of California, Irvine, CA, 92697, USA.}
\affil[2]{Department of Mechanical and Aerospace Engineering, University of California, Irvine, CA 92697, USA.}
\affil[3]{Department of Materials Science and Engineering, University of California, Irvine, CA 92697, USA.}
\affil[4]{Department of Civil and Environmental Engineering, University of California, Irvine, CA 92697, USA.}

\begin{document}
\include{pythonlisting}
    \pagenumbering{arabic}
    \sloppy
    \maketitle
    \noindent \textbf{Abstract}\\
Achieving desired mechanical properties in additive manufacturing requires many experiments and a well-defined design framework becomes crucial in reducing trials and conserving resources. Here, we propose a methodology embracing the synergy between high-throughput (HT) experimentation and hierarchical machine learning (ML) to unveil the complex relationships between a large set of process parameters in Laser Powder Bed Fusion (LPBF) and selected mechanical properties (tensile strength and ductility). The HT method envisions the fabrication of small samples for rapid automated hardness and porosity characterization, and a smaller set of tensile specimens for more labor-intensive direct measurement of yield strength and ductility. The ML approach is based on a sequential application of Gaussian processes (GPs) where the correlations between process parameters and hardness/porosity are first learnt and subsequently adopted by the GPs that relate strength and ductility to process parameters. Finally, an optimization scheme is devised that leverages these GPs to identify the  processing parameters that maximize combinations of strength and ductility. By founding the learning on larger “easy-to-collect” and smaller “labor-intensive” data, we reduce the reliance on expensive characterization and enable exploration of a large processing space. Our approach is material-agnostic and herein we demonstrate its application on 17-4PH stainless steel. 

    \section{Introduction} \label{sec intro}
The demand for lightweight and high-performance materials with intricate geometries has fueled the growth of additive manufacturing (AM) technologies \cite{DebRoy2018AdditiveProperties, Bajaj2020SteelsProperties, Tolosa2010StudyStrategies, King2015LaserChallenges}. Among the various AM processes, laser powder bed fusion (LPBF) has emerged as a leading method for the production of metallic components with exceptional mechanical properties and design flexibility \cite{Tolosa2010StudyStrategies, Yap2015ReviewApplications, Fields2024MicrostructuralFusion, Fields2024InvestigationProperties}. In LPBF, a thin layer of the feedstock powder is first placed on a substrate and is selectively melted by a laser beam at locations specified by a computer-aided design (CAD) model. The substrate is then lowered by one layer thickness and the processes of powder deposition and melting are repeated until the desired part is fabricated \cite{Tolosa2010StudyStrategies, King2015LaserChallenges}. Due to the highly localized melting and strong temperature gradients and cooling rates, LPBF can provide non-equilibrium microstructures that are not achievable via conventional techniques \cite{Wang2017AdditivelyDuctility, Vunnam2019EffectMelting, Voisin2021NewPowder-bed-fusion}. The layer-by-layer nature of the LPBF technique enables fabrication of near-net-shape complex parts with minimal need for post-processing and machining \cite{DebRoy2018AdditiveProperties, Bajaj2020SteelsProperties, Saeidi2015HardenedMelting}. 

The properties of parts built by LPBF heavily depend on the process parameters as they control the material's structural features at multiple length scales. The mechanical properties of highest technological importance, including yield strength, strain hardening behavior, fracture toughness and ductility (or strain to failure), are strongly influenced by features at the 10-100 micron scale (e.g., porosity, defects and inclusions) as well as micro/nano-structural features including phase evolution, precipitate formation and distribution, grain structure (size and texture), and solidification structures (dendrites, etc...) \cite{DebRoy2018AdditiveProperties, Chowdhury2022LaserModelling, Sofinowski2021Layer-wise316L, Fields2024MicrostructuralFusion, Fields2024InvestigationProperties, Moyle2022EvidenceSteel, Haines2023ExperimentalSteel, Moyle2022OnDesign}. 
These structural features are programmed by not only the complex dynamics of the melt pool including selective evaporation of alloying elements, turbulence, and convective motions (e.g., Marangoni flows), but also the cooling rates from the molten state and repeated thermal cycling as adjacent sections of the part are built \cite{Khairallah2016LaserZones, Kizhakkinan2023LaserReview, Agrawal2022PredictiveSolutions, Fields2024MicrostructuralFusion, Sabooni2021LaserProperties, Haines2023ExperimentalSteel}. These phenomena are all controlled by dozens of processing parameters, the most important of which include laser power, scan speed, shape and size of the laser spot, layer thickness, hatch spacing, and printing strategy \cite{Agrawal2020High-throughputSteel, Zhao2020AOptimization, Weaver2021DemonstrationCharacterization}. 

Fully elucidating the complex material-specific relationships between processing parameters, microstructural evolution and mechanical properties in LPBF, with a level of accuracy that enables optimization of processing parameters, remains a formidable challenge \cite{Gullane2022ProcessFusion}. While computational approaches have certainly helped \cite{King2015LaserChallenges}, the wide range of the involved length and time scales necessitate the adoption of multiple computational models. For example, microstructural evolution is best captured by combinations of molecular dynamics \cite{Jamil2022MolecularProcess, Peng2023APowders}, phase field, and CALPHAD techniques \cite{Liu2019InsightManufacturing, Fields2024InvestigationProperties}, whereas heat and mass transfer in the melt pool rely on computational fluid dynamics, and thermal stress evolution is generally modeled with the finite elements method \cite{Khairallah2016LaserZones, Chowdhury2022LaserModelling}. Tying all these approaches together in a full multi-physics package is a formidable task. At the same time, the number of processing parameters is too large to optimize them via brute-force experimentation. Consequently, predictive modeling of process-property relations in LPBF has traditionally relied on domain knowledge and trial-and-error methods \cite{Agrawal2022High-throughputMaterials}. 

To date, the prevailing approach to process parameter optimization has been to distill a small number of physical quantities that embed the most critical parameters, and experimentally scan them to identify the optima. Volumetric Energy Density (VED) is one of the most popular of such feature, which is defined as the laser power divided by the product of scan speed, hatch spacing (i.e. the distance between adjacent scan lines), and layer thickness \cite{Chowdhury2022LaserModelling}. Non-dimensional versions of VED have been introduced as an attempt to make this quantity material-independent \cite{Ion1992DiagramsProcessing,Thomas2016NormalisedAlloys}. VED has been correlated with ``print quality'' for multiple materials: low values of VED generally result in lack-of-fusion (LOF) porosity, whereas high values can cause keyhole porosity. Hence, optimizing the processing parameters for printed parts generally involves fabricating small samples over a range of VED values, characterizing their porosity (via optical microscopy, CT scanning, and/or Archimedes density measurements) to identify the optimal VED range for printing, and finally adopting combinations of laser power, scan speed, hatch spacing and layer thickness that result in this optimal VED. While this approach has been successfully demonstrated for multiple materials, two key challenges remain: $(1)$ while high values of porosity are certainly deleterious to mechanical properties, other microstructural features mentioned above may play an equally significant role; $(2)$ while VED has an appealing physical interpretation (i.e., the amount of energy embedded in a volume of material through the printing process), there is no guarantee that it fully characterizes the effect of all process parameters on the material structure and properties. 

To address some of these challenges, high-throughput (HT) techniques have been developed, which involve creating large arrays of samples with variations in composition or process parameters, followed by testing and screening to identify the conditions that yield optimal properties \cite{Agrawal2022High-throughputMaterials}. Compared to traditional approaches, HT techniques offer faster experimentation with reduced systematic errors and enhanced data reliability \cite{Agrawal2022High-throughputMaterials}. In the context of AM and especially LPBF, some efforts with HT approaches have been made to correlate process parameters, microstructure, and properties of fabricated materials \cite{Agrawal2020High-throughputSteel, Zhao2020AOptimization, Weaver2021DemonstrationCharacterization, Huang2021HighOptimization}.
Research in this area has involved automated tensile property characterization \cite{Huang2021HighOptimization, Heckman2020AutomatedSensitivity, Salzbrenner2017High-throughputSteel}, alloy design by using feedstock materials with varying chemical composition \cite{Zhang2024HighEvolution}, high-rate part fabrication \cite{Gullane2022ProcessFusion, Clare2021InterlacedGeometries}, and sample design and characterization to link process parameters with material properties \cite{Agrawal2020High-throughputSteel, Agrawal2022High-throughputSteel, Weaver2021DemonstrationCharacterization, Huang2021HighOptimization}.
Even in the HT context, exploring a broad processing space is quite time-consuming, and hence most efforts primarily focused on varying only laser power and scan speed over relatively small ranges and few different conditions \cite{Agrawal2022High-throughputSteel, Huang2021HighOptimization}, thus lacking insights into the effect of other process parameters or large parameter variations.
Additionally, existing maps for parameter selection are either non-predictive \cite{Thomas2016NormalisedAlloys} or validated only for specific materials \cite{Agrawal2022High-throughputSteel, Huang2021HighOptimization}.

As experimental data collection techniques advance, approaches based on machine learning (ML) are increasingly used to build data-driven process-property relations \cite{Wang2022Data-drivenDirections, Ko2023AManufacturing}. However, as printing and microstructurally/mechanically characterizing even a few dozen of samples produces stochastic data \cite{Agrawal2020High-throughputSteel, Huang2021HighOptimization, Heckman2020AutomatedSensitivity, Salzbrenner2017High-throughputSteel} and is very expensive and time-consuming, the resulting ML models are not readily applicable to constructing process parameter relationships and process design optimization in LPBF. 

\begin{figure} [!t]
    \centering
    \includegraphics[width=1\linewidth]{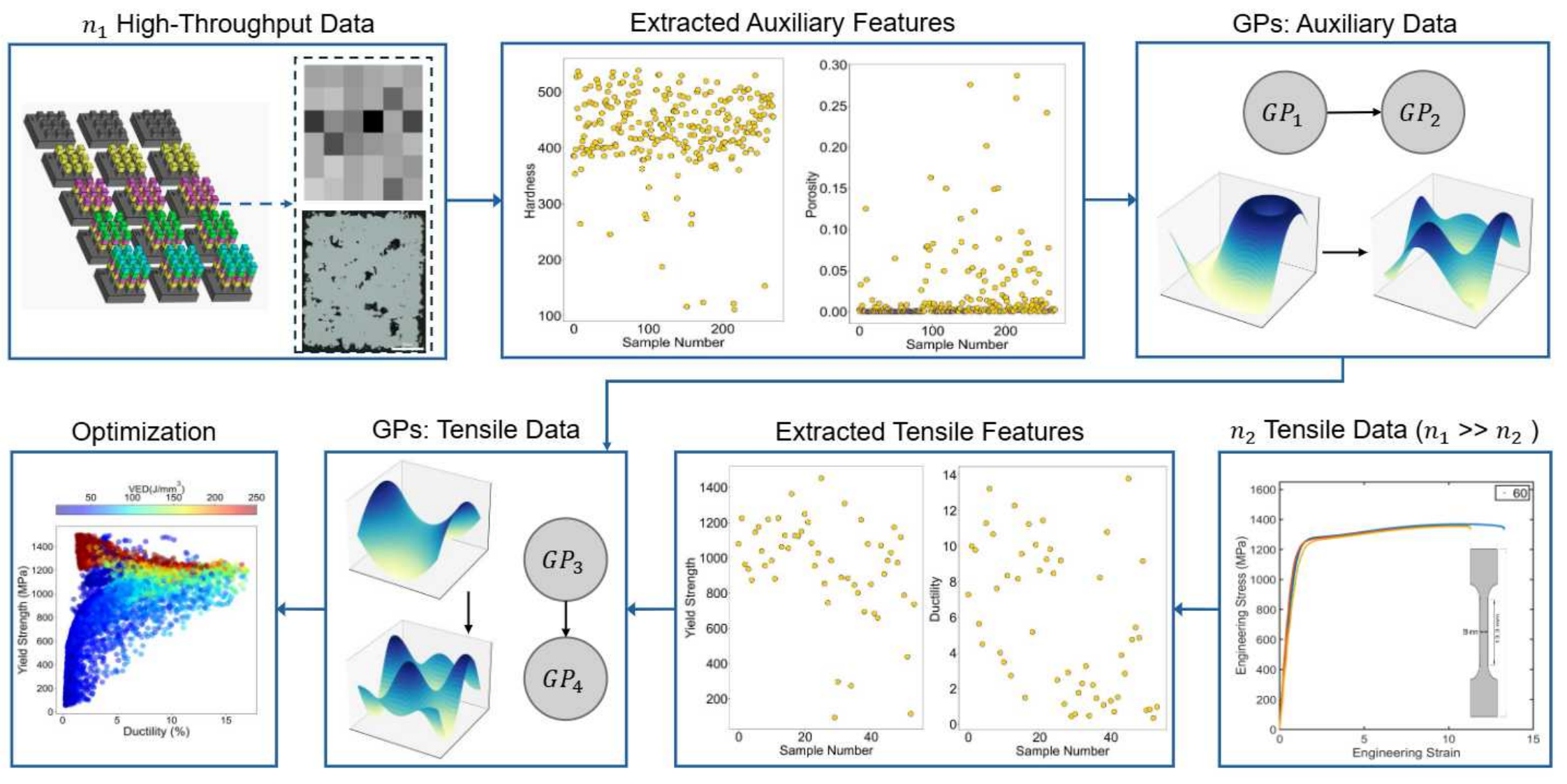}
    \caption{\textbf{Schematic flowchart of the proposed framework:} High-throughput experimental approaches are coupled with hierarchical learning based on Gaussian processes to design the process parameters that optimize the combination of tensile strength and ductility.}
    \label{fig main flowchart}
\end{figure}

In this work, we develop a novel ML approach coupled with HT printing and characterization investigations to optimize a wide range of LPBF processing parameters (laser power $p$, laser scan speed $v$, hatch spacing $h$, powder layer thickness $l$, and scan rotation between layers $sr$) to achieve maximum combinations of material yield strength (\YS) and ductility (\D). The framework is depicted in \Cref{fig main flowchart}. As fabrication and testing of multiple dog bone specimens required for direct measurements of \YS~and \D~as a function of processing parameters is extremely costly and time-consuming, we propose a two-step experimental process to generate suitable training data for ML: $(1)$ We print a large set of small cuboid samples spanning the entire processing parameter space and rapidly characterize their surface properties (here chosen as hardness and porosity); as hardness maps are obtained on each cuboid, this results in robust statistics. $(2)$ We print a relatively small number of tensile dog bone specimens over a sub-set of the parameter space and test them via uniaxial loading to directly extract their \YS~and \D. Similarly, we use a two-step ML approach based on Gaussian processes (GPs) \cite{LMGP, gp+,oune2021latent} to learn the complex correlations between processing parameters and \YS~and \D: first, two GPs are built to relate process parameters to hardness and porosity; subsequently, two more GPs are trained that leverage the first two GPs as well as the additional tensile data.  Finally, an optimization scheme is devised to identify the process parameters that maximize combinations of strength and ductility.

The rationale behind our approach is that the information embedded in easy-to-measure surface properties (hardness maps and porosity) is correlated with the properties of interest (strength and ductility). While such correlations are not surprising, their functional form is unknown to us and can only be quantified in a data-driven manner via ML models. In our case, the complexity of these correlations is very high because our two datasets based on surface and tensile measurements are $(1)$ affected by LPBF process parameters in different ways, $(2)$ based on samples whose shapes and sizes significantly differ (cuboid vs tensile coupons), and $(3)$ unbalanced since we have far more data from cuboids which are easy to manufacture and test. We demonstrate that by leveraging these unknown correlations within our framework it is possible to explore a very large parameter space and build ML models that can identify processing settings that produce samples with desirable mechanical properties.

While the proposed approach is material agnostic, here we demonstrate it with 17-4 PH stainless steel (SS), a precipitation-hardened alloy with diverse industrial applications requiring high strength and corrosion resistance \cite{Fields2024MicrostructuralFusion, Vunnam2019EffectMelting, Sabooni2021LaserProperties}. While in the conventional wrought form this steel is fully martensitic, multiple studies clearly indicate that the microstructure of LPBF-processed \alloy~is very complex, often consisting of combinations of martensite, ferrite, and occasionally residual austenite \cite{Fields2024MicrostructuralFusion, Vunnam2019EffectMelting, Li2022HomogenizationFusion, Moyle2022OnDesign, Sabooni2021LaserProperties, Yeon2022NormalizingFusion, Lebrun2015EffectSteel, Murr2012MicrostructuresMelting}. The microstructure is strongly related to both the selective evaporation of ferrite- and austenite-stabilizing alloying elements and the local thermal cycles experienced during the LPBF process which are highly sensitive to the processing parameters \cite{Fields2024MicrostructuralFusion, Vunnam2019EffectMelting, Sabooni2021LaserProperties, Yeon2022NormalizingFusion, Haines2023ExperimentalSteel}. Strong evidence also exists that clearly relates microstructure to tensile properties in this material \cite{Fields2024MicrostructuralFusion}, making \alloy~the perfect alloy to demonstrate the power of the proposed approach.


    \section{Materials and Methods} \label{sec Method}

\subsection{Design and Manufacturing} \label{subsec design}
Nitrogen atomized \alloy\ powder with a particle size range of $15-50~ \mu m$  (Carpenter Additive, USA) was used as the feedstock material. Printing was carried out using an SLM Solutions $125$HL printer, featuring a Yb-fiber laser with a maximum output of $400$ W and a beam diameter of $80~ \mu m$. The build chamber operated in a $99.99\%$ N$_2$ atmosphere where the build plate was preheated to $200~$°C, and a ``Stripe" scan strategy was used. Fill contour and border scans were not used to obtain a uniform microstructure across thin wall samples. These machine settings were used for the manufacturing of both cuboids and tensile coupons.

A design of experiments (DOE) methodology based on the Sobol sequence was employed to generate $270$ LPBF process parameter combinations, with laser power ($p$) varying in the range $80-400$ W, laser scan speed ($v$) in the range $150-1500$ mm/s, powder layer thickness ($l$) in the range $20-75~$ µm, hatch spacing ($h$) in the range $70-120$ µm, and two possible scan rotation ($sr$) values of $67$ or $90$ degrees. The $270$ combinations produced VED values ranging approximately from $10$ to $1000$ J/mm$^3$ and are presented in \Cref{tab: parameter sets}. These process settings were used to create $270$ cube-shaped $2\times2\times2$ mm samples which were numbered as $1, \cdots, 270$ based on their corresponding process parameter combination. 

\begin{figure} [!h]
    \centering
    \includegraphics[width=0.9\linewidth]{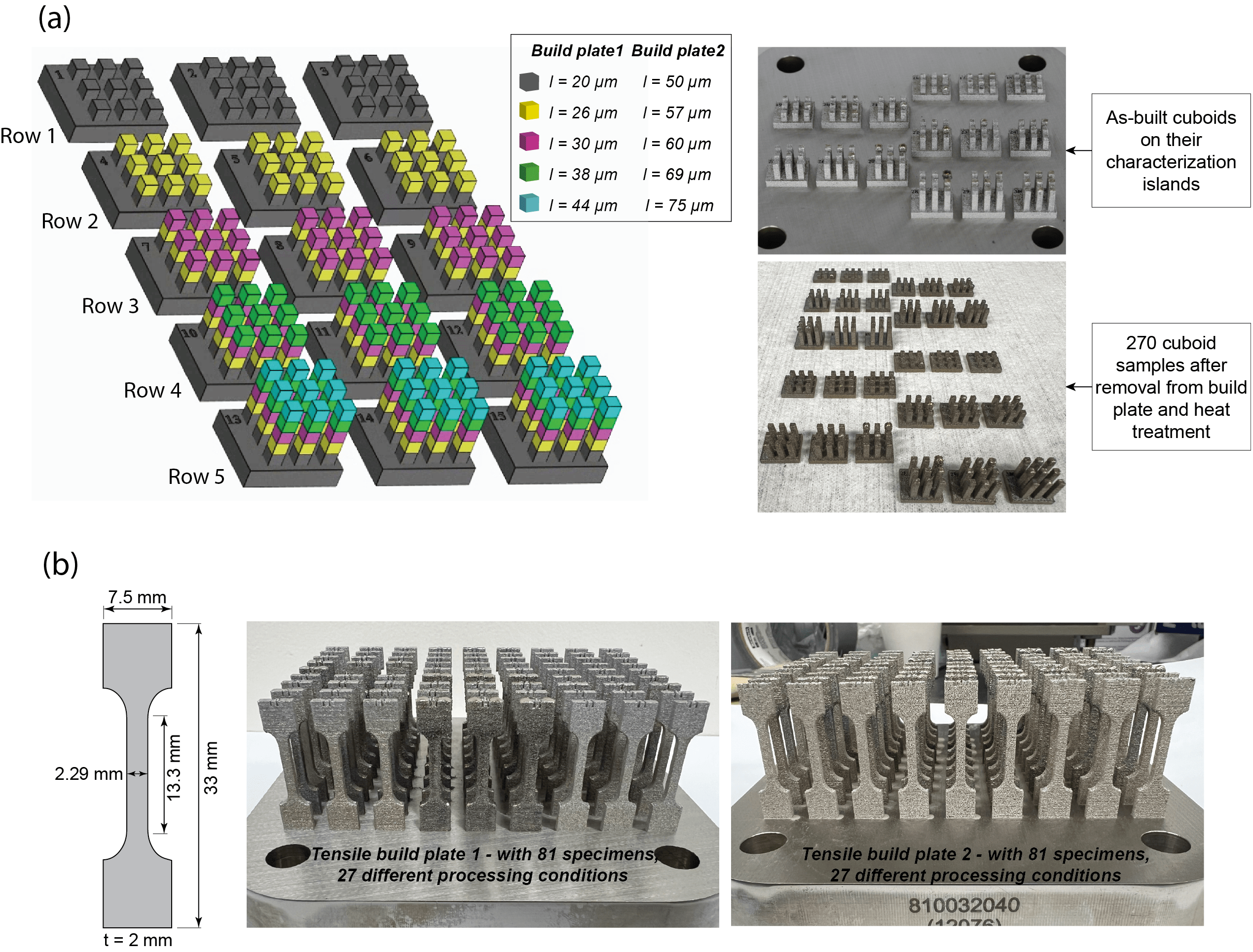}
    \caption{\textbf{Designed experimental setups:} \textbf{(a)} Schematic of the HT-compatible build design for cuboids with different layer thicknesses, along with the real LPBF generated cuboids, and \textbf{(b)} Tensile specimen dimensions and LPBF printed samples.}
    \label{fig sample design}
\end{figure}

To enable HT manufacturing and surface characterization of cuboids, a unique island-based setup was designed. This setup allows for the manufacturing of samples with several varying layer thicknesses on one build plate, while facilitating sample removal via electro-discharge machining (EDM) and subsequent polishing and characterization of multiple samples concurrently. $15$ islands of size $14\times 12 \times 3$ mm, each containing $9$ cuboids with different processing conditions, were positioned in the chamber and numbered as shown in \Cref{fig sample design}(a). This process was repeated a second time on a separate build plate to obtain islands $16$ through $30$. The layer thickness values in the $270$ combinations were projected to $10$ unique levels, i.e., $l \in \braces{20, 26, 30, 38, 44, 50, 57, 60, 69, 75}$ µm, where the first $5$ values were used for islands $1$ through $15$ and the rest for islands $16$ through $30$. 

The island-based process is schematically illustrated in \Cref{fig sample design}(a), where each color represents a specific layer thickness. First, the cuboids in grey were printed at a layer thickness of $20$ µm where printing was concluded once an overall height of $2$ mm was attained. 
Then, as shown via the yellow cuboids, printing was continued with a layer thickness of $26$ µm to add another $2$ mm of material to rows $2$ through $5$. 
This process was repeated for the three remaining layer thickness values. Hence, the cuboids shared the same layer thickness value if they were on islands that were positioned in the same row in \Cref{fig sample design}(a), but the other four process parameters changed across all cuboids, see \Cref{tab: parameter sets} for processing parameters used for cuboids on each island. The cuboids on the second build plate were printed in a similar manner, except that the layer thickness values started from $50$ µm and continued to $75$ µm. It is noted that using this approach even higher numbers of layer thicknesses can be printed on the same build plate, based on the manufacturing needs and part size limits.

Tensile specimens were printed in the vertical direction, meaning the build direction was parallel to the tensile loading direction. A subset of $270$ process parameter combinations, specifically the $54$ combinations that correspond to layer thickness values of $30$ and $60$ µm, were used for manufacturing of the tensile specimens (see the highlighted rows in \Cref{tab: parameter sets}). These two layer thickness values were selected because they fall in the range of the layer thickness values for which cuboids were built. 
For each $54$ process parameter combination, three replicas were printed to assess the variability of tensile properties under the same process parameters. The dimensions of the tensile specimens and images of the printed tensile coupons are provided in \Cref{fig sample design}(b).

LPBF printed samples were removed from the build plate via wire electro-discharge-machining (EDM). All cuboids and tensile coupons were heat treated upon their removal from the build plate to increase their hardness and yield strength \cite{Lashgari2023HeatSteel}. The samples were directly aged (with no prior solutionization step)  at $482$ °C for $1$ h in a Nabertherm B$400$ furnace in an ambient atmosphere with a heating rate of $10$ °C min$^{-1}$, and then air quenched. 

\subsection{Microstructural and Mechanical Characterization} \label{subsec characterization}
For microstructural characterization and hardness testing, islands of cuboids were embedded in epoxy/resin mounts, ground, and polished via standard procedures for stainless steels down to $1$ µm with diamond polishing suspensions (MetaDi, Buehler). Finally, samples were chemically-mechanically polished with 0.05 µm Alumina suspension (MasterPrep Alumina, Buehler). Etching was done using Waterless Kalling's, also known as Kalling's No.$2$ Reagent (ES Laboratory, LLC) to distinguish between phases in \alloy. Samples were submerged in the etchant for approximately $24$ s,  immediately rinsed, sonicated in water for $1$ min, and air dried. Microstructural analyses of $270$ cuboids were performed using an Olympus DSX$10$-UZH Digital Optical Microscope. Polished and etched surfaces of cuboids were imaged for analysis of the defects (pores and cracks) and microstructure phases.

A robust porosity measurement approach was developed to extract the porosity content of each sample based on the OM images obtained from the as-polished surfaces. To refine the images, mitigate noise effects, and enhance clarity and quality in this process, preprocessing techniques such as blurring and cropping were first used. Then, based on the distribution of the pixel values shown in \Cref{fig: pixel_dist}, a threshold of $75$ was selected to distinguish pores, i.e., pixels whose brightness is below $75$ were classified as pores. Subsequently, porosity was computed as the ratio of pore area to the total image area.
More details on image processing and porosity calculations are provided in \Cref{sec: porosity_measurement}.

Vickers microindentation hardness mapping was selected as the rapid HT mechanical property characterization. Instrumented indentation is well-suited for HT testing because it can quickly measure location-specific mechanical responses, has automation capability, and only requires a flat polished surface for testing \cite{Weaver2021DemonstrationCharacterization}. 
The Vickers microindentation hardness test utilizes a calibrated machine to apply a square-based pyramidal-shaped diamond indenter with face angles of 136° to the material's surface. Test forces range from $1$ to $1000$ gf ($9.8\times10^{-3}$ to $9.8$ N) and the resulting indentation impressions (diagonals) are measured using a light microscope upon the load removal. Then, the Vickers hardness (kgf/mm$^2$) is determined as $HV = 1.8544~F/d^2$ where $F$ is force (kgf) and $d$ is the mean diagonal length of the indentations (mm).
Vickers hardness measurements were obtained via a Buehler Wilson VH$3300$ automated indenter for the $270$ cuboids. Each cuboid underwent $36$ measurements spaced $280$ µm apart, using a $0.5$ kgf load and a $10$ s hold time for each indent. The median of these $36$ measurements was recorded as the hardness value for each sample. Median was used rather than the mean to reduce the effect of outliers. 

Tensile tests were performed on an Instron $5985$ load frame equipped with a $250$ kN load cell. Tensile specimens were tested with their surfaces in the as-printed condition following the heat treatment, without any surface machining or polishing prior to the test. Each specimen was marked with two white circular fiducial marks setting the gauge length limits for strain tracking. An AVE$2663-901$ video extensometer with a Fujinon HF$16$HA-$1$S lens was used to track the strain of the gauge section. Tests were conducted according to ASTM E$8$ standards at a quasi-static strain rate of $0.001$ s$^{-1}$. The obtained stress-strain curves were assessed to extract the 0.2$\%$ offset yield strength (\YS), strain to failure (\D), and ultimate tensile strength (\UTS).
These three parameters were extracted using a built-in software in the Instron $5985$.

To emphasize the impact of this HT approach to fabrication and testing, we estimate that the entire set of prints employed approximately $6$~kg of material and took approximately $14$~hr. For reference, if we had printed exclusively tensile dog-bones with the same number of different processing parameters ($270$), with three repetitions per condition, we would have needed to print 810 samples, on at least 10 platform. We estimate that this would have required approximately $22$~kg of material, and a print time of about $48$ hrs. Hence, our approach resulted in a $~3.5X$ reduction in both material cost and print time (a very conservative estimate, as we are not factoring in the EDM time as well as LPBF preparation time, which scales linearly with number of platforms), as well as a $5X$ reduction in tensile testing time.

\subsection{Hierarchical Learning via Gaussian Processes} \label{sec ML_method}
\begin{figure} [!t]
    \centering
    \includegraphics[width=0.8\linewidth]{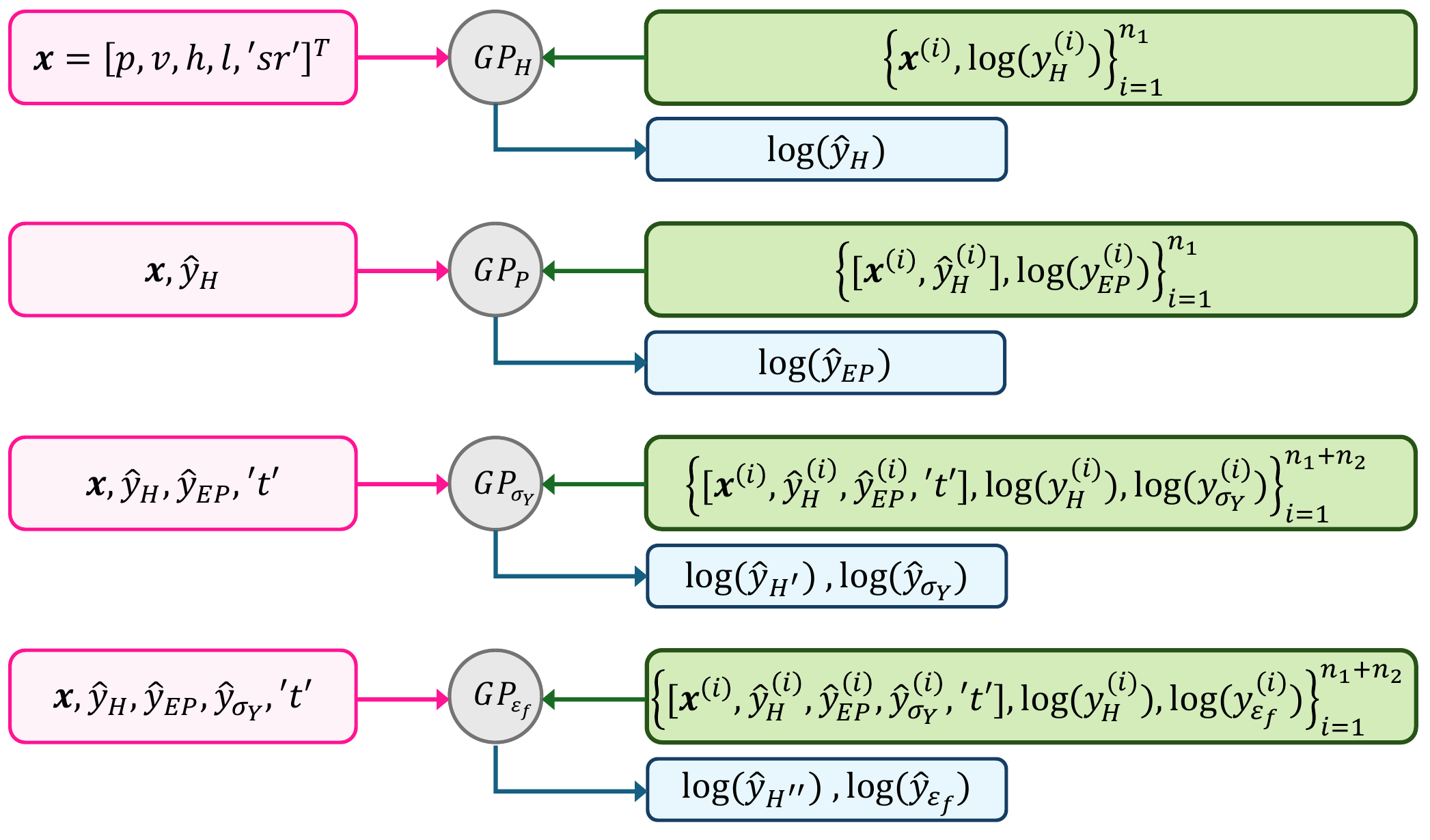}
    \caption{\textbf{Proposed hierarchical learning framework:} Input-output spaces and model structures vary depending on the characteristics of each property.}
    \label{fig: hierarchical_learning}
\end{figure}

We propose an ML framework to leverage \textit{auxiliary features} obtained from hardness maps and OM images of cuboids towards the overarching goal of predicting the mechanical properties of tensile coupons. As illustrated in \Cref{fig: hierarchical_learning}, our framework has a hierarchical nature, where we first learn to predict the auxiliary features as functions of process parameters and then use their \textit{predicted} values as additional inputs in the GPs that estimate tensile properties. Our ML framework is designed based on two critical assumptions: (i) the auxiliary features and tensile properties are naturally related (since they are material characteristics), and depend on the same set of process parameters; (ii) hardness and porosity are measured within an HT scheme which leads to far more samples on the auxiliary features than on \YS~and \D~which are obtained via tensile tests. Hence, we do not learn the dependence of these two sets of properties on process parameters with a single ML model because such a model would have to be trained on the combined dataset, which is highly imbalanced. Such a dataset would cause the ML model to primarily focus on hardness and porosity, while our ultimate goal is to predict tensile properties as functions of process parameters. To mitigate this issue while leveraging the relation between the two sets of properties, we build predictive models for the auxiliary features first, and subsequently use them in ML models that predict \YS~and \D. 

The relation between the auxiliary features and tensile properties is \textit{hidden}, i.e., we rely on data to model this relation as there is no analytic physics-based formula that can relate hardness and porosity of cuboids to tensile properties of dog-bone specimens in LPBF. The complexity of this hidden relation is especially high in our case due to the size-effect: while the auxiliary features are obtained from surface of small cuboid samples, \YS and \D~are based on tensile tests performed on much larger coupons. To unveil these complex hidden relations, we use GPs in our framework because they can distill highly complex input-output relations even from small datasets. 

Below, we first review some technical details on GPs in \Cref{sec: back_GP} and then elaborate on how they are used in our hierarchical learning framework in \Cref{sec hl framework}. Our framework is implemented via the open-source Python package \gp \cite{gp+}.

\subsubsection{Emulation and Data Fusion via Gaussian Processes (GPs)} \label{sec: back_GP}
GPs are probabilistic models that assume the training data follows a multivariate normal distribution. They are defined by parametric mean and covariance functions whose parameters can be systematically optimized via maximum likelihood estimation (MLE) or maximum a posteriori (MAP). Once the parameters are estimated, closed-form conditional distribution formulas are used for probabilistic prediction \cite{schulz2018tutorial,deringer2021gaussian,trinchero2020combining,wan2014analytical,hamdi2015gaussian}. GPs are particularly suited for our application as they $(1)$ can naturally handle noise, $(2)$ do not rely on big data, and $(3)$ can efficiently learn complex input-output relations  \cite{williams2006gaussian,kachabi2024markov,CMAME_SANAZ}. 

In this work, we design the mean and covariance functions of the GPs to seamlessly $(1)$ handle the categorical variable $sr$ in the process parameter space, and $(2)$ enable data fusion or multi-fidelity (MF) modeling, which refers to the process of jointly learning from multiple datasets that share some mutual information. MF modeling is essential in this work as we aim to combine hardness/porosity and tensile data together to leverage their connection and, in turn, reduce the reliance on expensive tensile data. 

As detailed in \Cref{sec: back_GP_appendix}, traditional GPs cannot handle categorical variables directly, as they are not naturally endowed with a distance metric. 
To address this limitation, \gp~ first uses the user-defined fixed function $f_{\pi}(\tb)$ to transform the categorical variable $\tb$~into the quantitative representation $\pib_t$. To reduce the dimensionality of $\pib_t$ while capturing its effects on the response, $\pib_t$ is then passed through the parametric embedding function $f_h(\pib_t; \thetab_h)$ with parameters $\thetab_h$. Since the outputs of $f_h(\pib_t; \thetab_h)$ are low-dimensional and quantitative, they can be easily integrated with the mean and covariance functions of the GP. Upon this integration, all model parameters are estimated via MAP. 

\begin{figure} [!h]
    \centering
    \includegraphics[width=1\linewidth]{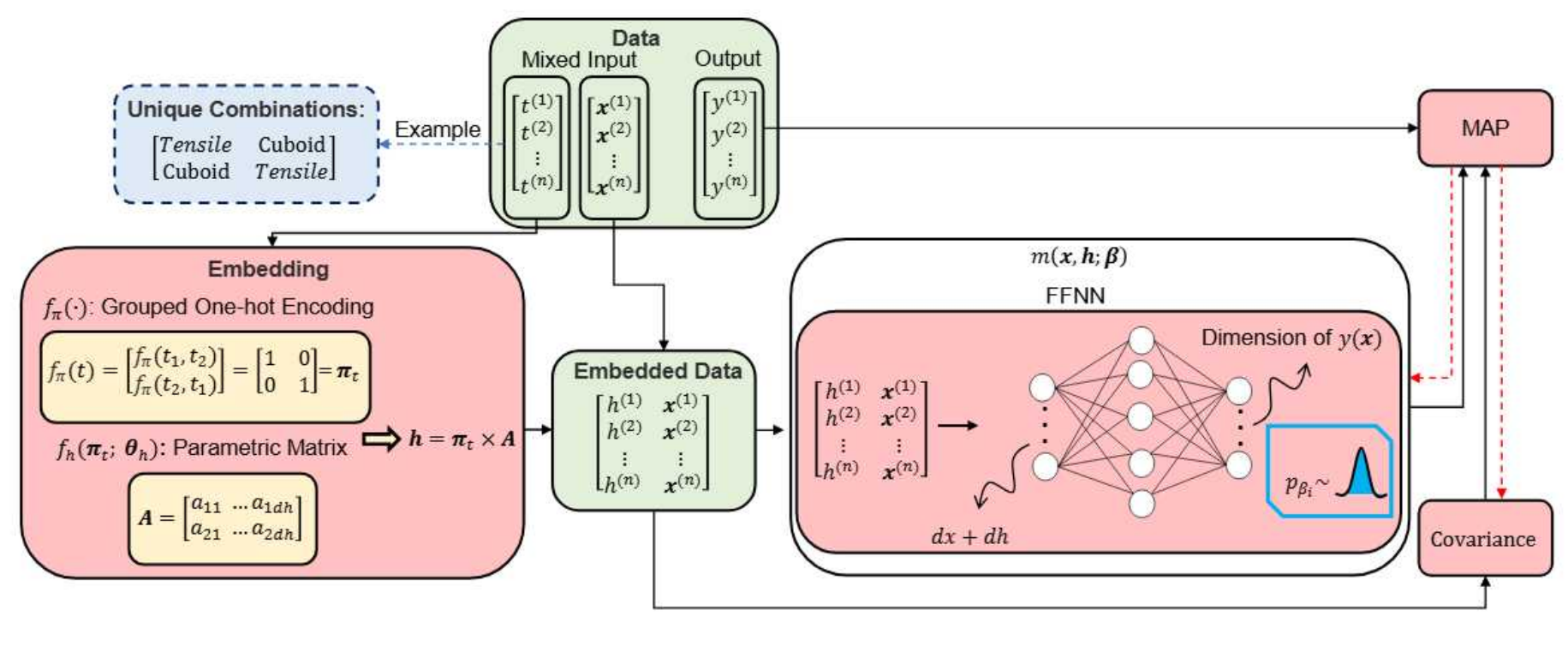}
    \caption{\textbf{Data Fusion via GPs:} To fuse the hardness/porosity with tensile data using \gp, we add a two-level categorical variable $t = \{Cuboid, Tensile\}$ (source indicator) to the input space. We use grouped one-hot encoding and matrix multiplication to convert $t$ to its low-dimensional quantitative representation $\hb$. Then, these mapped values ($\hb$) are concatenated with the quantitative input features and fed into the mean and covariance functions. To capture more complex relations in the data, we use a FFNN as a mean function and all the model parameters are estimated via MAP.}
    \label{fig: GP}
\end{figure}

\gp~ enables MF modeling by simply augmenting the input space with an additional categorical feature which merely indicates the source of a data point, see \Cref{fig: GP} where $t=\braces{Cuboid, Tensile}$ is the added two-level categorical variable, $\xb$ are the numerical inputs, and $y$ denotes the output. This approach provides the option of having a mean function \(m(\xb, \hb; \betab)\) with parameters $\betab$ that is either dependent or independent of the data source (\Cref{fig: GP} shows the former case as $t$ affects the mean function through $\hb$).
As detailed in \Cref{sec hl framework}, MF modeling is used to fuse the hardness/porosity and tensile datasets. To this end, the two-level categorical variable $t=\braces{Cuboid, Tensile}$ is added to the input space and is converted to $\pib_t$ and then $\hb$ via grouped one-hot encoding and matrix multiplication, respectively. Additionally, the mean function is modeled via a feed-forward fully-connected neural network (FFNN) and all model parameters are estimated via MAP.

\subsubsection{Hierarchical Learning} \label{sec hl framework}
As explained earlier, the GPs for the auxiliary features are built first, and then used by the GPs that predict tensile properties. To devise the hierarchy among the four variables, we calculate the Pearson correlation coefficient \cite{cohen2009pearson} between any pair of variables. This coefficient provides values between $-1$ and $1$, where the sign and magnitude indicate the direction (direct or inverse) and strength of the correlation, respectively. While these correlations are linear, they provide a good starting point for our nonlinear analyses.

The results are enumerated in \Cref{tab: correlations} and indicate that there is a very high linear correlation between \UTS~and \YS. Hence, we exclude \UTS~from the analysis since $(1)$ the same procedure for predicting \YS~can be repeated for \UTS, and $(2)$ \YS~and \UTS~are not competing properties in our case (unlike \D~and \YS) and hence process optimization is unaffected.
The values in \Cref{tab: correlations} also indicate that hardness and porosity are more correlated with \YS~ than \D; motivating us to learn the former first and then use it for learning the latter.

\begin{table}[!h]
    \centering
    \begin{tabular}{cccccc}
    \cline{2-6}
    & Hardness & Porosity & \text{\YS }& \UTS & \text{\D}                  \\ \cline{2-6} 
        \multicolumn{1}{c|}{Hardness}  & $1$    & $-0.56$    & $0.84$ & $0.78$ & \multicolumn{1}{c}{$0.26$}  \\
        \multicolumn{1}{c|}{Porosity}  & $-0.56$    & $1$        & $-0.79$ & $-0.80$ & \multicolumn{1}{c}{$-0.45$} \\
        \multicolumn{1}{c|}{\text{\YS}}     & $0.84$     & $-0.79$    & 1 & $0.99$    & \multicolumn{1}{c}{$0.53$}  \\
        \multicolumn{1}{c|}{\UTS}     & $0.78$     & $-0.80$    & $0.99$ & 1    & \multicolumn{1}{c}{$0.62$}  \\
        \multicolumn{1}{c|}{\text{\D}} & $0.26$     & $-0.4$5    & $0.53$ & $0.62$ & \multicolumn{1}{c}{1}     \\ \hline
    \end{tabular}
    \caption{\textbf{Pearson coefficients of correlation among properties:} Negative values show an inverse relationship, where one property tends to decrease as the other increases. The absolute values represent the strength of the correlation, with values closer to $1$ indicating a stronger relationship between the properties.}
    \label{tab: correlations}
\end{table}

Our framework is illustrated in \Cref{fig: hierarchical_learning} where pink and blue boxes represent the input and output spaces of each model, respectively, green boxes indicate the training data, $n_1$ and $n_2$ refer to the number of cuboid and tensile data, respectively, and $\boldsymbol{x}$ denotes the process parameters. 
The subscript $s$ on the output variables distinguishes different properties, i.e., $s \in \{H, EP, \text{\YS}, \text{\D}\}$, with $y_s$ and $\hat{y}_s$ representing the experimental measurements and predicted values, respectively. 

As shown in \Cref{fig: hierarchical_learning}, we train $GP_H$ first to predict hardness as a function of $\xb$. The reason for initializing the modeling sequence with hardness is that porosity prediction depends on it. As shown in \Cref{subsec hardness porosity}, the majority of porosity values in our dataset are very small which makes it difficult to relate their variations to process parameters. For example, the difference in the porosity of samples $14$ and $21$ in \Cref{tab: parameter sets} is only $10^{-5}$, which is a very small number that might be interpreted as noise by an ML model. 
To address this issue while maximally leveraging the (negative) correlation between hardness and porosity, we use the estimated hardness values not only as an additional input variable, but also for engineering an output feature. Since the scales of porosity and hardness are substantially different, we design this engineered porosity as $\text{\hh} \times exp(\text{\yp})$ and denote it by \EP~to ensure $(1)$ the variations of hardness do not dominate those of porosity, and $(2)$ small porosity values are not rounded to zero. With this engineered feature, the difference between samples $14$ and $21$ becomes $\approx 42$ which, in turn, makes it much easier to link the variations of $\xb$ and \EP. We denote the model that predicts \EP~as a function of process parameters and predicted hardness by $GP_{EP}$.

Once $GP_H$ and $GP_{EP}$ are trained, they are used for predictive modeling of tensile properties. For learning \YS, we leverage its relation with cuboids' hardness/porosity by $(1)$ augmenting the process parameters with the predicted hardness and the predicted engineered porosity feature, and $(2)$ fusing the two datasets obtained from tensile and cuboid samples (only hardness). We denote the resulting model by $GP_{\text{\YS}}$ and highlight that the second step increases the number of parameters in $GP_{\text{\YS}}$, as it must predict both \YS~ and a dummy hardness value\footnote{Since the predicted hardness of this model is never used, we treat it as a dummy variable which merely enables the data fusion.}, rather than only \YS. However, despite the increase in the number of parameters, data fusion increases the dataset size from $54$ to $324$ ($54$ tensile samples plus $270$ cuboid samples), which justifies the increase in the number of parameters.

The final step involves learning ductility, which is more challenging due to its lower correlation with other properties and its significant inherent variability. To address these challenges, we augment the process parameters with all previously modeled properties (i.e., predicted hardness, predicted engineered porosity, and predicted tensile strength), and also fuse the two datasets obtained from tensile and cuboid samples. We denote the resulting model with $GP_{\text{\D}}$ which, similar to $GP_{\text{\YS}}$, benefits from data fusion and predicts a dummy hardness variable. 

We note that $'t'$ in the input space of $GP_{\text{\YS}}$ and $GP_{\text{\D}}$ is a categorical variable that differentiates data types and enables data fusion as detailed in \Cref{sec: back_GP_appendix}. This allows $GP_{\text{\YS}}$ and $GP_{\text{\D}}$ to predict two distinct values based on the value assigned to $'t'$. 
For example, $'t' \in \{'H', 'YS'\}$ in the case of $GP_{\text{\YS}}$ which predicts \hy~for samples with $'t'='YS'$ and a dummy hardness for $'t'='H'$. 
It is also noted that we design the mean function of each GP in \Cref{fig: hierarchical_learning} based on the dataset size and complexity. These details are included in \Cref{sec results} where we also evaluate the advantages of augmenting tensile models with auxiliary features and using data fusion.

\subsection{Accuracy Assessment} \label{sec: accuracy_testing}
To evaluate the accuracy of the GP models, $5$-fold cross-validation (CV) is used by partitioning the data into $5$ folds and then iterating over them. In each iteration, one fold serves as the validation set while the remaining $4$ folds are used for training \cite{CV1,CV2,CV3}. 
We measure the emulation accuracy in each iteration via mean squared error (MSE):
\begin{equation} 
    \begin{split}
        MSE=\sqrt{\frac{1}{n_{\text{test}}} \sum_{i=1}^{n_{\text{test}}}({y_s}^{(i)}-{\hat{y}_s}^{(i)})^2}
    \end{split}
    \label{eq mse}
\end{equation}
\noindent where ${y_s}^{(i)}=y_s({\boldsymbol{u}}^{(i)})$ and ${{\hat{y}_s}^{(i)}}={\hat{y}_s}{({\boldsymbol{u}}^{(i)})}$ denote, respectively, the median of the experimentally measured value and the prediction for sample ${\boldsymbol{u}}^{(i)}$. Subscript $s$ is defined \Cref{sec hl framework} and distinguishes different properties.

It is highlighted that since the experimental data are noisy, the MSE in \Cref{eq mse} cannot be smaller than the (unknown) variance of the noise, which is caused by a number of factors such as measurement errors and manufacturing variability. Hence, to have a baseline for assessing the magnitude of the MSE that a GP model provides on a held-out fold, we compare it to the noise variance. Since the noise variance is unknown, we estimate it for each property by fitting a GP to the entire data and reporting the estimated nugget parameter, see \Cref{eq: gp-mean-var-vec-noise}. These estimated noise variances are $\widehat\tau^2_H =5 \times 10^{-3},~\widehat\tau^2_{EP} =2 \times 10^{-4},~\widehat\tau^2_{\text{\YS}} =99 \times 10^{-4},~\widehat\tau^2_{\text{\D}}=0.05$

For evaluation, we also report the coefficient of determination (R$^2$) which, unlike MSE, is calculated based on the entire data via:
\begin{equation}
    \begin{split}
        R_{s}^2=1-\frac{\sum_{i=1}^n\left({y_s}^{(i)}-{\hat{y}_s}^{(i)}\right)^2}{\sum_{i=1}^n\left({y_s}^{(i)}-\bar{y}_s\right)^2}
    \end{split}
\end{equation}
\noindent where $\bar{y}_s$ indicates the average of (the median of) property $s$ over the $n$ samples. An R$_{s}^2$ close to $1$ indicates that the trained model can adequately explain the variability of the output with respect to the inputs. The obtained R$_{s}^2$ values by the GPs are R$_{H}^2=0.93$, R$_{EP}^2=0.98$, R$_{\text{\YS}}^2=0.94$, and R$_{\text{\D}}^2=0.68$.

    \section{Results and Discussions} \label{sec results}

We provide a detailed analysis of the properties of the cuboids and tensile coupons in \Cref{subsec hardness porosity,subsec tensile}, respectively. As detailed in \Cref{sec ML_method}, we link these properties to process parameters via four GP emulators that are trained hierarchically. We illustrate the prediction accuracy of these GPs using the metrics defined in \Cref{sec: accuracy_testing}. These emulators are used in \Cref{subsec optimization} to optimize the combination of yield strength and ductility of a tensile sample which is then built and tested to assess the effectiveness of our framework.

Throughout, we use the median of the properties to mitigate the effects of outliers. Specifically, hardness refers to the median value observed in the hardness map of a cuboid. For the tensile coupons, \YS, $\sigma_U$, and \D~(strain to failure/ ductility) refer to the respective median values observed in the three tensile test repetitions that are carried out for each process parameter combination. 
For brevity, we drop the word median when referring to these properties.

\subsection{Hardness and Porosity} \label{subsec hardness porosity}

\begin{figure}[!b]
  \centering
    \includegraphics[width=\textwidth]{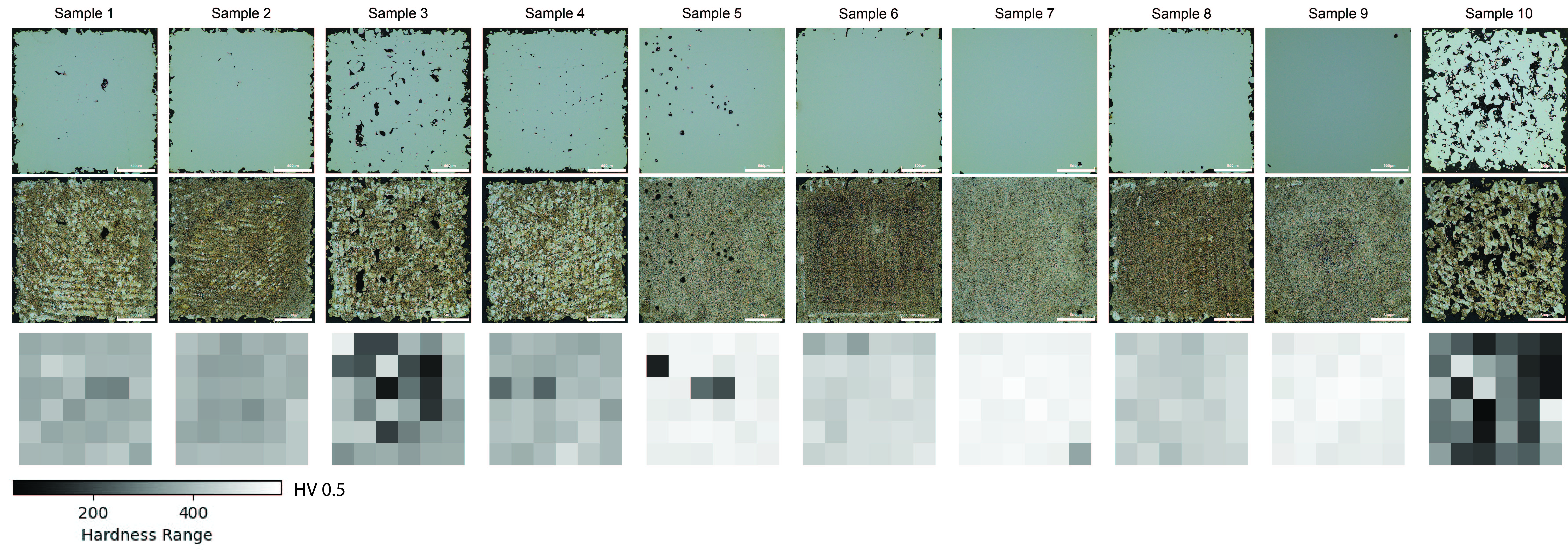}
    \vspace{-5mm}
    \caption{\textbf{Representative microstructural images and hardness maps of cuboids:} The top row includes the as-polished surfaces that illustrate the concentration of defects. The second and third row show, respectively, phase formation and hardness maps. This figure highlights the large impact of processing conditions on the microstructure, defect content, and property of the samples.}
    \label{fig: h_p_rep}
\end{figure}

\begin{figure}[!t]
  \centering
    \includegraphics[width=0.8\textwidth]{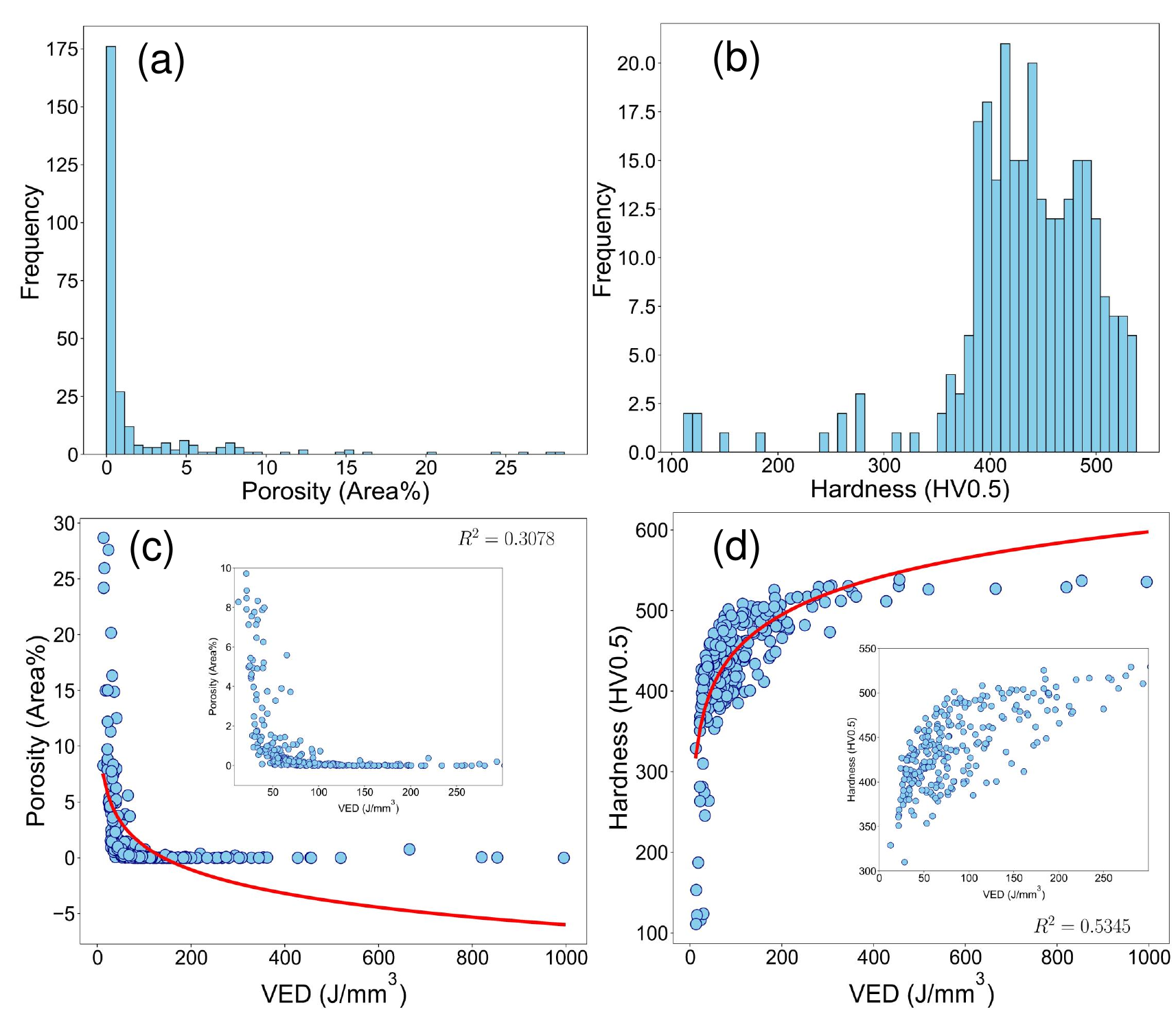}
    \vspace{-5mm}
    \caption{\textbf{Porosity and hardness:} (a) and (b) show porosity and hardness distributions among the $270$ cuboids, highlighting the significant dependency of these properties on the laser process parameters. (c) and (d) show variations of porosity and hardness vs VED, along with the corresponding R$^2$ of their fitted curves. The insets show the magnified image of a smaller region of energy density and property.}
    \label{fig: H_P_VED}
\end{figure}

Representative microstructural images of the polished and etched surfaces alongside the hardness maps of the $270$ cuboid samples are shown in \Cref{fig: h_p_rep} (see \Cref{fig: OM_H_all} for the complete set of images). 
The top row in \Cref{fig: h_p_rep} corresponds to the as-polished images, which show the spatial distributions of microdefects. 
Images in the second row are obtained from the etched surfaces and illustrate the formed phases, indicating that the samples have either a fully martensitic or a duplex ferritic/martensitic microstructure \cite{Fields2024MicrostructuralFusion}.
The bottom row in \Cref{fig: h_p_rep} includes representative hardness maps from the cuboids where darker pixels correspond to lower hardness. 
Overall, these images indicate that the relative content of phases, defect concentrations, and hardness significantly vary depending on the processing parameters, especially since we explore a wide range for each parameter. \Cref{fig: H_P_VED}(a) and \Cref{fig: H_P_VED}(b) provide the histograms of porosity and hardness of the $270$ cuboids which further demonstrate the strong dependency of these properties on processing parameters.

VED is widely used in the literature to correlate mechanical properties and defect content to the LPBF process parameters \cite{Fields2024MicrostructuralFusion, Huang2021HighOptimization, Tan2018MicrostructuralDirections, Bang2021EffectComposition}. To assess the predictive power of VED in our case, we plot the porosity and hardness of the cuboids against their corresponding VEDs, see \Cref{fig: H_P_VED}(c) and \Cref{fig: H_P_VED}(d).
As shown in \Cref{fig: H_P_VED}(c), porosity initially decreases at a high rate as energy density increases and then it plateaus once the energy density exceeds roughly $100$ J/mm$^3$ due to the removal of LOF porosity. 
Expectedly, we observe an opposite trend for hardness in \Cref{fig: H_P_VED}(d), where it increases and then nearly plateaus as VED increases. The initial rapid change in hardness is due to the increase in the structure's density and reduction in the LOF porosity. For equally dense structures in \Cref{fig: H_P_VED}(d), we observe that hardness shows a roughly increasing trend as VED exceeds $\sim70$ J/mm$^3$ and then plateaus at $\sim400$ J/mm$^3$ VED. We attribute this trend to the higher evaporation of Cr from the melt pool with higher VED \cite{Fields2024MicrostructuralFusion}. As Cr is a ferrite stabilizer, this evaporation promotes formation of austenite, which transforms into martensite upon the fast cooling during the LPBF process. Importantly, however, the increase in hardness with VED does not apply to all the processing conditions: the inset in \Cref{fig: H_P_VED}(d) shows a broad set of data where a specific relationship between hardness and VED cannot be inferred. 

As evidenced by \Cref{fig: H_P_VED}(c) and \Cref{fig: H_P_VED}(d), it is notable that samples with identical energy densities can display significant variations in porosity and hardness. This result clearly implies that VED cannot fully capture the impact of processing parameters on hardness for 17-4PH steel.
To further elaborate this point, we leverage curve fitting to regress the data via a wide range of analytic functions\footnote{We use polynomials, $log(\cdot), exp(\cdot),$ and combinations thereof.} and report the one with the highest R$^2$ in \Cref{fig: H_P_VED}(c) and \Cref{fig: H_P_VED}(d). Since the VED of the majority of the samples is below $300$ J/mm$^3$, the fitted curves prioritize these regions and hence fail to provide accurate predictions at high VEDs, where much smaller property variations are observed. Hence, even the best fitted curves provide low R$^2$ values, supporting the conclusion that VED alone is insufficient to accurately predict porosity or hardness.

The wide variations of porosity and hardness, coupled with the insufficiency of VED in accurately linking them to the process parameters, motivate the use of ML models.
As explained in Section \ref{sec hl framework}, we first relate hardness to the process parameters via a GP model and then use this model while building another GP that predicts the porosity of cuboids as a function of process parameters and the predicted hardness. To assess features' importance and potentially reduce the input dimensionality, we calculate Sobol's sensitivity indices (SI) based on the GP model that predicts hardness. As detailed in \Cref{sec Sobole}, these indices are based on variance analysis and quantify the importance of a variable on the output either solely on its own (main SI) or including its interactions via other variables (total SI). These SI indices are enumerated in \Cref{table: SA_hardness}, and indicate that the most important process parameters are laser power and speed, followed by layer thickness and hatch spacing. Scan rotation has a negligible effect on hardness, and hence can be excluded from the ensuing studies. As the difference between main and total SIs indicates the effect of variable interactions on the output, we deduce from the numbers in \Cref{table: SA_hardness} that the four process parameters affect hardness in a non-trivial manner. For instance, we observe that the effect of hatch spacing (or layer thickness) on hardness increases by about five times as other process parameters vary. Additionally, since the main SIs are not zero (especially in the case of laser power and speed), we can once again conclude that a single variable such as VED (which only consists of variable interactions) cannot explain the variability of the output as the inputs change.


\begin{table}[]
\centering
\begin{tabular}{cc|ccccc}
 &          & \multicolumn{5}{c}{Processing Parameters}  \\ \hline
\multicolumn{1}{c|}{Property} & Metric   & Laser Power & Laser Speed & Layer Thickness & Hatch Spacing & \multicolumn{1}{c}{Scan Rotation} \\ \hline
\multicolumn{1}{c|}{\multirow{2}{*}{\text{\yh}}} & Main SI  &  $0.392$       & $0.302$      & $0.011$ & $0.008$   & $0.000$  \\
\multicolumn{1}{c|}{}  & Total SI & $0.685$       & $0.565$      & $0.072$ & $0.039$   & $0.000$  \\ \hline
\multicolumn{1}{c|}{\multirow{2}{*}{\text{\EP}}} & Main SI  &$0.174$       & $0.138$      & $0.077$ & $0.012$   & $0.009$  \\
\multicolumn{1}{c|}{}& Total SI & $0.485$       & $0.366$      & $0.155$ & $0.199$   & $0.019$  \\ \hline
\end{tabular}
\caption{\textbf{Main and total Sobol sensitivity indices:} These indices quantify the effect of a variable on the response either solely on its own (main SI) or including that variable's interactions via other variables (total SI).}
    \label{table: SA_hardness}
\end{table}

Upon excluding scan rotation from the inputs, we conduct $5$-fold CV via GPs and report the corresponding MSEs in the top row of \Cref{tab: CV_h_p} (the CV plots are included in \Cref{fig: CVs}). By comparing the MSEs to $\widehat\tau^2_H = 5\times 10^{-3}$,
we observe that these values are quite close. This observation, in conjunction with the large R$_{H}^2=0.93$, indicates that GPs can effectively predict hardness as a function of process parameters. Additionally, the consistent MSE values across different folds show the robustness of the trained model. 
We note that in two cases the reported MSEs are smaller than noise variance which is due to the fact that the latter is also ``estimated".
We obtain our final model for hardness prediction by fitting a GP to the entire $270$ samples and henceforth denote it by $GP_H$.


\begin{table}[!b]
\centering
\begin{tabular}{c|ccccc}
                    & \multicolumn{5}{c}{MSE}                                        \\ \hline
Property            & Fold 1 & Fold 2 & Fold 3 & Fold 4 & \multicolumn{1}{c}{Fold 5} \\ \hline
\text{\yh}            & $0.014$      & $0.007$      & $0.003$      & $0.033$ & $0.003$   \\ \hline
\text{\EP} & $6 \times 10^{-4}$   & $6 \times 10^{-4}$     & $1 \times 10^{-4}$     & $11 \times 10^{-4}$     & $3 \times 10^{-4}$                          \\ \hline
\end{tabular}
\caption{\textbf{Accuracy assessment based on MSEs:} The errors are reported for $5$-fold CV for both hardness and porosity prediction.}
    \label{tab: CV_h_p}
\end{table}

As shown in \Cref{fig: hierarchical_learning}, we use the estimated hardness values from $GP_H$ to augment the input and output (i.e., the engineered porosity) spaces of the porosity data. 
The sensitivity indices in the bottom row of \Cref{table: SA_hardness} illustrate similar trends as in hardness; laser power and speed are the most important process parameters, followed by layer thickness, hatch spacing, and scan rotation. 
The different main SIs represent varying degrees of each process parameter's effect on engineered porosity, which cannot be captured by VED. Additionally, the low main and total SIs for scan rotation indicate its small impact on the engineered porosity feature. 
Considering this negligible effect, we exclude it from the input space.

We conduct $5$-fold CV via GPs trained on the first four process parameters concatenated with the predicted hardness values from $GP_H$. The calculated MSEs are presented in \Cref{tab: CV_h_p} and plots depicting its performance on different folds are available in \Cref{fig: CVs}.
Similar to hardness, the close similarity between MSEs and $\widehat\tau^2_{EP}=2 \times 10^{-4}$, high R$_{EP}^2=0.98$, and the consistency of MSEs among different folds illustrate the effectiveness and robustness of GPs to predict \EP~(and hence ductility).
We denote the final model for predicting \EP~as $GP_{EP}$ and obtain it by fitting a GP to the entire data which is augmented with the predicted hardness values.

\subsection{Tensile Properties} \label{subsec tensile}
\begin{figure}[!h]
    \centering
    \includegraphics [width=1\linewidth]{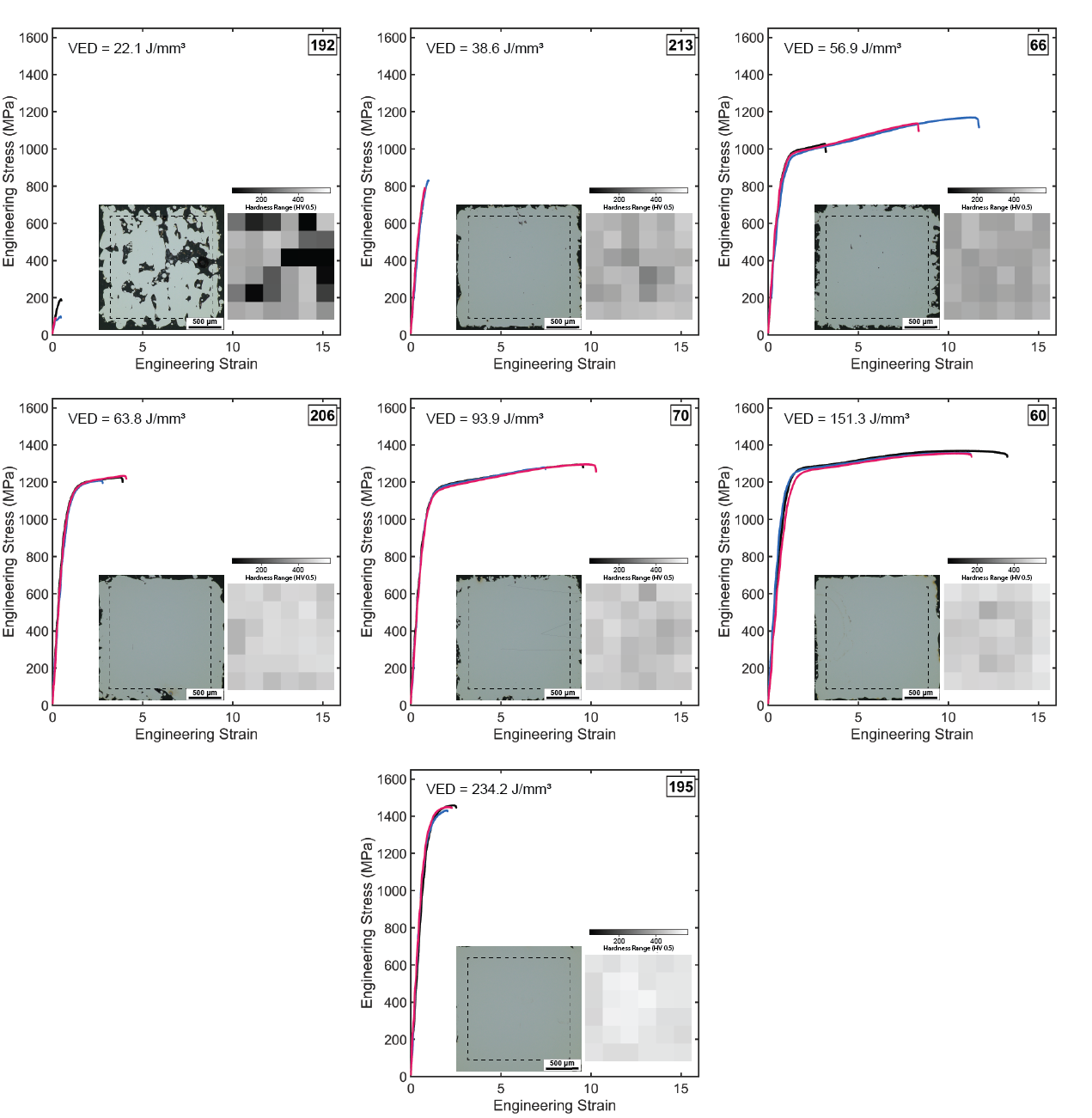}
    \caption{\textbf{Representative tensile properties:} Representative stress-strain curves for dog-bone samples printed with  7 different processing parameters (3 nominally identical samples were tested for each condition). Insets display porosity and hardness maps from cuboids printed with the same processing conditions. The dashed lines indicate the area mapped for hardness. The number on the right corner shows the sample number (see  \Cref{tab: tensile properties} for processing conditions).}
    \label{fig representative_tenisle}
\end{figure}

\begin{figure}[!t]
  \centering
    \includegraphics[width=\textwidth]{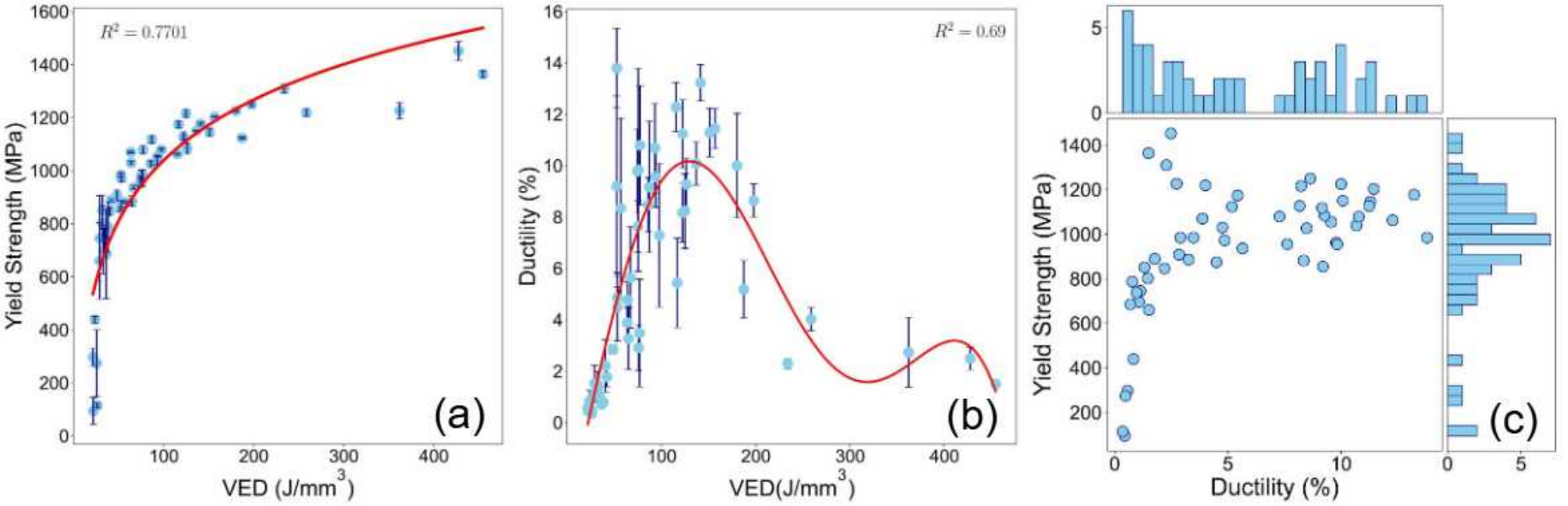}
    \vspace{-3mm}
    \caption{\textbf{Yield strength and ductility:} (a) and (b) present variations of \YS~and \D~vs VED, along with the R$^2$ of their corresponding fitted curves. (c) illustrates the trade-off between \YS~and \D, with the axes showing their distributions among the $54$ tensile specimens, highlighting the significant dependency of these properties on the laser process parameters.}
    \label{fig: hist_ved_tensile}
\end{figure}

We build tensile specimens based on $54$ process parameter combinations which are also used to manufacture cuboids $55-81$ and $190-216$ as enumerated in \Cref{tab: parameter sets}. We provide representative stress-strain curves in \Cref{fig representative_tenisle} and refer the reader to \Cref{fig: stress-strain curves} for the complete set. In \Cref{fig representative_tenisle} each plot also contains the hardness map and OM image of the corresponding cuboid. We extract \YS~, \UTS, and \D~from the $54$ stress-strain curves, and present the values in \Cref{tab: tensile properties}. Given the strong correlation between \UTS~and \YS~observed in \Cref{tab: correlations}, we only analyze \YS.
The histograms and plots of the variation of \YS~and \D~against each other and VED is shown in \Cref{fig: hist_ved_tensile}, which indicates that these two properties $(1)$ vary quite substantially across the different samples, and $(2)$ are nonlinearly related, whereby \D\ first increases but then decreases as \YS~increases.
As indicated by the whiskers in \Cref{fig: hist_ved_tensile}(c) and \Cref{fig: hist_ved_tensile}(d), \D~ shows higher stochasticity than \YS, as commonly reported for AM metals \cite{Heckman2020AutomatedSensitivity, Salzbrenner2017High-throughputSteel}.

The plots in \Cref{fig representative_tenisle} clearly show the diversity of samples in terms of tensile responses and microstructures. For instance, cuboid and tensile specimens $192$ are built with a very small VED of $22.1$ J/mm$^3$, which does not provide sufficient energy for proper material consolidation. Such a VED results in a cuboid with many large pores, a small hardness value of $368.7$ HV$0.5$, and a scattered hardness map depending on the indentation location, see also \Cref{fig: H_P_VED}. This processing condition also produces weak tensile specimens whose yield stress and ductility are roughly below $100$ MPa and $1\%$, respectively.
Further increase in VED to the level that LOF porosity is lowered to its plateau value increases the hardness, \YS, and \D~as seen in process combinations $213$, $66$, and $206$. VEDs above $200$ J/mm$^3$ can result in embrittlement of the manufactured samples, leading to considerably high strength but low ductility values. For example, processing condition $195$ in \Cref{fig representative_tenisle} with a very high VED of $234.2$ J/mm$^3$ has a \YS~of $1,310$ MPa and \D~of 2.3$\%$, respectively. Overall, we observe that samples printed with VEDs of roughly $100$ to $200$ J/mm$^3$ exhibit enhanced ductility and tensile strength (see specimens $70$ and $60$ in \Cref{fig representative_tenisle}) compared to samples printed with VED values outside this range. However, there is no clear relationship between VED variations and \YS~and \D~, either within or outside this range. 
To better demonstrate this point, we consider process parameters $66$ and $206$ in \Cref{fig representative_tenisle}, which result in similar VEDs but the corresponding cuboids and tensile coupons have dissimilar microstructures, surface properties, and stress-strain curves. Specifically, we observe that tensile specimen $66$ produces stress-strain curves with higher ductility and larger stochasticity but lower strength.
Such property variations in \alloy~happen mainly due to the changes in martensite/austenite/ferrite ratios (see \Cref{fig: OM_H_all}), porosity percentage, precipitation, and elemental evaporation and chemical composition with changes in the processing conditions \cite{Fields2024MicrostructuralFusion, Lebrun2015EffectSteel, Sabooni2021LaserProperties, Li2022HomogenizationFusion, Moyle2022OnDesign, Murr2012MicrostructuresMelting}.

Considering the complex relationship between processing parameters and tensile properties and variabilities in the measured properties, we arrive at a similar conclusion to \Cref{subsec hardness porosity} in that VED fails to fully explain the dependency of \D\ and \YS~on the process parameters and can only achieve modest R$^2$ values, as observed from the fitted regression models shown in \Cref{fig: hist_ved_tensile}(a) and \Cref{fig: hist_ved_tensile}(b). Hence, we rely on our GP-based hierarchical framework to distill these relations from the datasets. As demonstrated in \Cref{fig: hist_ved_tensile}(a), \YS~is less stochastic than \D, and more correlated with hardness and porosity. Therefore, we begin by linking \YS~to process parameters and predicted hardness and porosity (see \Cref{fig: hierarchical_learning}). Then, we use the trained GP, along with the GPs built-in \Cref{subsec hardness porosity}, to predict \D\ as a function of process parameters and previously predicted properties.

As explained in \Cref{sec hl framework}, we leverage the relation between \YS~of tensile coupons and cuboid properties by $(1)$ augmenting the process parameters with the predicted hardness and the engineered porosity, and $(2)$ fusing the two datasets from tensile and cuboid samples. The resulting model is denoted by $GP_{\sigma}$.
To assess the impact of these choices, we compare the performance of $GP_{\sigma}$ to two baseline GPs denoted by $GP_{\sigma'}$ and $GP_{\sigma''}$. $GP_{\sigma'}$ leverages neither of the above steps (i.e., it only uses the $54$ tensile samples to link \YS~to process parameters), while $GP_{\sigma''}$ only excludes the data fusion step.

\begin{table}[!b]
\centering
\begin{tabular}{cc|ccccc}
 &          & \multicolumn{5}{c}{Processing Parameters}  \\ \hline
\multicolumn{1}{c|}{Property} & Metric   & Laser Power & Laser Speed & Layer Thickness & Hatch Spacing & \multicolumn{1}{c}{Scan Rotation} \\ \hline
\multicolumn{1}{c|}{\multirow{2}{*}{\text{\yy}}} & Main SI  &  $0.260$       & $0.125$      & $0.093$ & $0.036$   & $0.003$  \\
\multicolumn{1}{c|}{}  & Total SI & $0.636$       & $0.400$      & $0.345$ & $0.308$   & $0.058$  \\ \hline
\multicolumn{1}{c|}{\multirow{2}{*}{\text{\yd}}} & Main SI  &$0.073$   & $0.108$   & $0.282$ & $0.044$   & $0.001$  \\
\multicolumn{1}{c|}{}& Total SI & $0.450$       & $0.368$      & $0.607$ & $0.259$   & $0.081$  \\ \hline
\end{tabular}
\caption{\textbf{Sobol sensitivity indices for yield strength and ductility:} Unlike in \Cref{tab: CV_h_p}, layer thickness and hatch spacing are quite important when predicting \YS~and \D~based on the process parameters.}
    \label{table: SA_tensile}
\end{table}

The SIs for \YS~are shown in the top row of \Cref{table: SA_tensile} and indicate that except scan rotation, all other process parameters affect \YS. This trend is similar to \Cref{tab: CV_h_p} but the interplay between the remaining four parameters is higher in this case (compare the total SIs across the two tables). 
We exclude scan rotation from the input space and conduct a $5$-fold CV study via $GP_{\sigma'}$. The MSEs are presented in the top row of \Cref{tab: CV_YS}, and the calculated noise variance and R$^2$ are $22\times 10^{-4}$ and $0.18$, respectively. 
The very low R$^2$ and substantial difference between MSEs and the estimated noise variance indicate the inability of $GP_{\sigma'}$ to accurately predict \YS. This expected result is due to the small dataset size and the high complexity of the relation between~\YS~and process parameters.


\begin{table}[!t]
\centering
\begin{tabular}{c|ccccc}
& \multicolumn{5}{c}{MSE}                                        \\ \hline
    Model  & Fold $1$ & Fold $2$ & Fold $3$ & Fold $4$ & \multicolumn{1}{c}{Fold $5$} \\ \hline
    $GP_{\sigma'}$& \multicolumn{1}{c|}{$0.294$}  & \multicolumn{1}{c|}{$0.059$}  & \multicolumn{1}{c|}{$0.475$}  & \multicolumn{1}{c|}{$0.681$}  & \multicolumn{1}{c}{$0.076$}  \\ \hline
    $GP_{\sigma''}$  &\multicolumn{1}{c|}{$0.085$}  & \multicolumn{1}{c|}{$0.046$}  & \multicolumn{1}{c|}{$0.342$}  & \multicolumn{1}{c|}{$0.463$}  & \multicolumn{1}{c}{$0.182$}  \\ \hline
    $GP_{\sigma}$  &  \multicolumn{1}{c|}{$0.096$}  & \multicolumn{1}{c|}{$0.016$}  & \multicolumn{1}{c|}{$0.296$}  & \multicolumn{1}{c|}{$0.0701$}  & \multicolumn{1}{c}{$0.166$}    \\ \hline
    $GP_{\varepsilon'}$& \multicolumn{1}{c|}{$0.194$}  & \multicolumn{1}{c|}{$0.238$}  & \multicolumn{1}{c|}{$0.451$}  & \multicolumn{1}{c|}{$0.416$}  & \multicolumn{1}{c}{$0.417$}  \\ \hline
    $GP_{\varepsilon}$  &\multicolumn{1}{c|}{$0.388$}  & \multicolumn{1}{c|}{$0.146$}  & \multicolumn{1}{c|}{$0.258$}  & \multicolumn{1}{c|}{$0.401$}  & \multicolumn{1}{c}{$0.334$}  \\ \hline
\end{tabular}
\caption{\textbf{Accuracy assessment based on MSEs:} The errors are reported for $5$-fold CV for both yield strength and ductility.}
\label{tab: CV_YS}
\end{table}


To improve the performance of the model, we augment the process parameters with the predicted hardness and porosity to train $GP_{\sigma''}$. The calculated MSEs of the corresponding $5$-fold CV study are presented in the middle row of \Cref{tab: CV_YS}. The R$^2$ and noise variance are calculated as $0.64$ and $0.07$, respectively. 
The improved R$^2$ as well as close MSEs and noise variance suggest that $GP_{\sigma''}$ leverages the predicted hardness and porosity to better explain the variability of \YS. Referring to \Cref{tab: correlations}, this enhancement is expected due to the high correlation observed among \YS, hardness, and porosity. 
To highlight the improvement level, we note that the minimum MSE obtained by $GP_{\sigma'}$ is approximately $26$ times larger than the estimated noise variance while in the case of $GP_{\sigma''}$ this ratio drops to $1.07$.
Despite the considerable improvement, feature augmentation cannot solely provide high accuracy due to the very small dataset size and hence $GP_{\sigma}$ also leverages data fusion.

We employ a source-dependent mean function in $GP_{\sigma}$ to capture the unique behaviors of each property while simultaneously modeling their underlying interdependencies.
We design the mean function based on an FFNN with three layers of two neurons to strike a balance between complexity and accuracy. Tangent hyperbolic (TH) is used as the activation function of each neuron and $20~\%$ dropout is added for regularization. 
The MSEs are presented in the third row of \Cref{tab: CV_YS} and the calculated noise variance and R$^2$ are $99\times 10^{-4}$ and $0.94$, respectively (see \Cref{fig: CVs} for a graphical error representation). 
The high R$^2$ value and similarity between MSEs and the estimated noise variance indicate that $GP_{\sigma}$ is effectively leveraging data fusion and feature augmentation. Henceforth, we denote this model as $GP_{\text{\YS}}$ and note that, expectedly, it provides higher errors compared to $GP_H$ and $GP_{EP}$ since \YS~has more variability, is a more complex property to predict, and there are fewer samples on it ($54$ vs $270$).

Following \Cref{fig: hierarchical_learning}, the final step involves learning \D, ~where process parameters are augmented with all of the previously predicted properties (\hh, \hp, \hy) and tensile data are fused with cuboid data. To evaluate the effectiveness of these two choices, we compare the performance of $GP_{\varepsilon}$ to a baseline GP denoted as $GP_{\varepsilon'}$ which does not incorporate these enhancements.

The Sobol SIs for \D~in the bottom row of \Cref{table: SA_tensile} highlight layer thickness and laser speed as the most important process parameters, followed by laser power, hatch spacing, and scan rotation. This ordering is different than the ones in \Cref{table: SA_hardness} which highlights the complexity of data fusion, i.e., our framework is expected to leverage hardness, porosity, and tensile strength features even though they are not affected by the process parameters in the same way as ductility.

\begin{figure}[!b]
  \centering
    \includegraphics[width=0.9\textwidth]{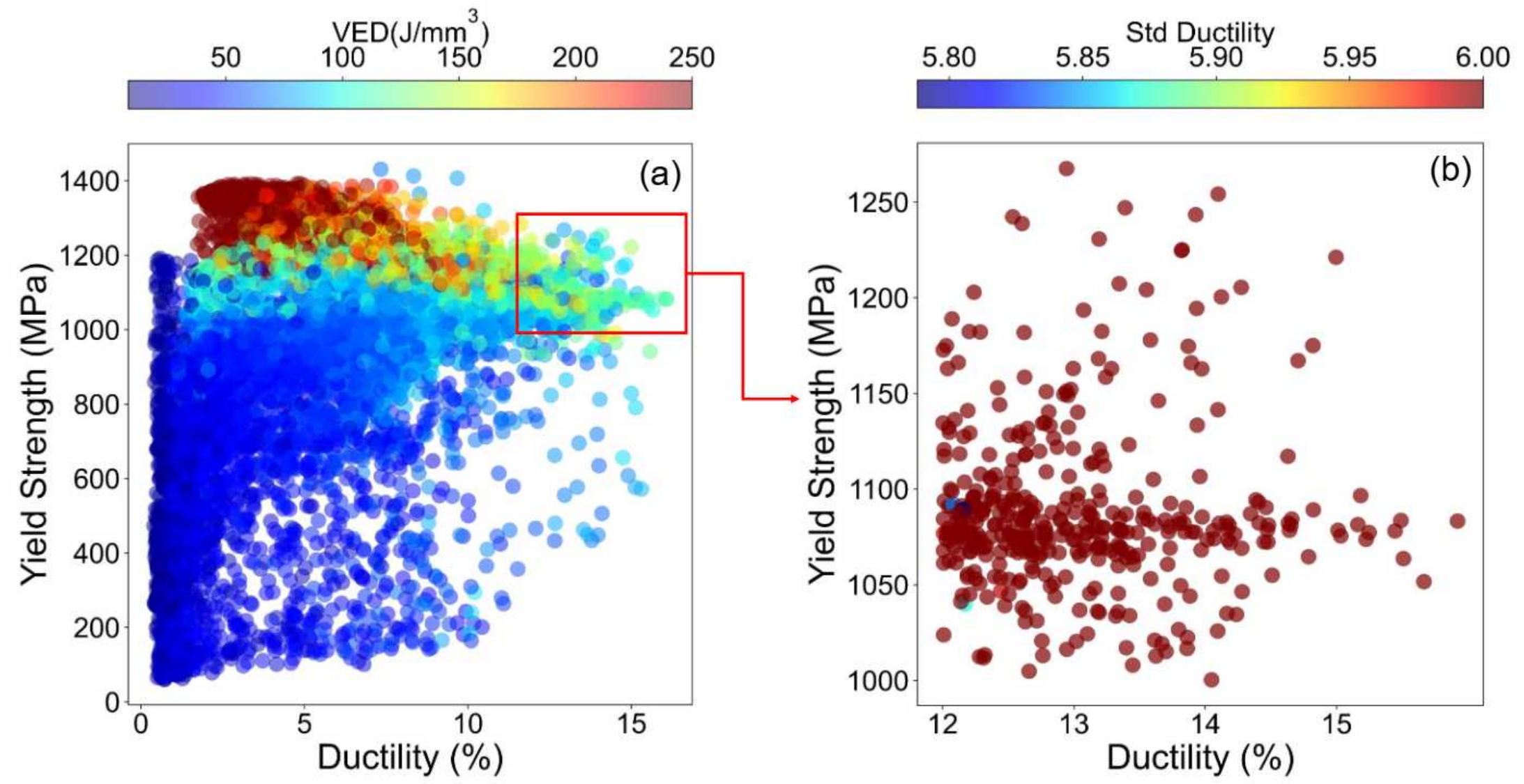}
    \vspace{-3mm}
    \caption{\textbf{Generalizability of $GP_{\varepsilon'}$:} \textbf{(a)} displays the $10,000$ generated samples color-coded based on their VED. In \textbf{(b)}, we narrow these samples to those within the optimal domain and color-code them based on the prediction uncertainty, highlighting the significant uncertainty of the model in this region.}
    \label{fig: generalizability_base1}
\end{figure}

The higher importance of layer thickness over laser power can be attributed to the fact that the GP model has only seen layer thickness values $\{30, 60\}$ and incurs some errors when predicting other values which, in turn, affects the SIs. 
Similar to previous cases, we exclude scan rotation from the input space (as its main and total SIs are small) and perform a $5$-fold CV using $GP_{\varepsilon'}$. 
The MSEs are reported in the forth row of \Cref{tab: CV_YS}, with estimated R$^2$ and noise variance values being $0.76$ and $0.16$, respectively. Considering the inherent challenges of learning ductility, these metrics indicate that the GPs perform relatively well but cannot generalize accurately. This issue is illustrated in \Cref{fig: generalizability_base1}, where we generate $10,000$ random process parameters and predict their corresponding \YS~and \D. We first color code the points based on their corresponding VED in \Cref{fig: generalizability_base1}(a) and then focus on the region with high \hy~and \hd—as shown in \Cref{fig: generalizability_base1}(b), which is color-coded based on the standard deviation of \hd. The predicted standard deviations are automatically provided by the GP (see \Cref{sec: back_GP_appendix}) and indicate that the corresponding predictions have large uncertainties and cannot be trusted for process optimization.

\begin{figure}[!t]
  \centering
    \includegraphics[width=\textwidth]{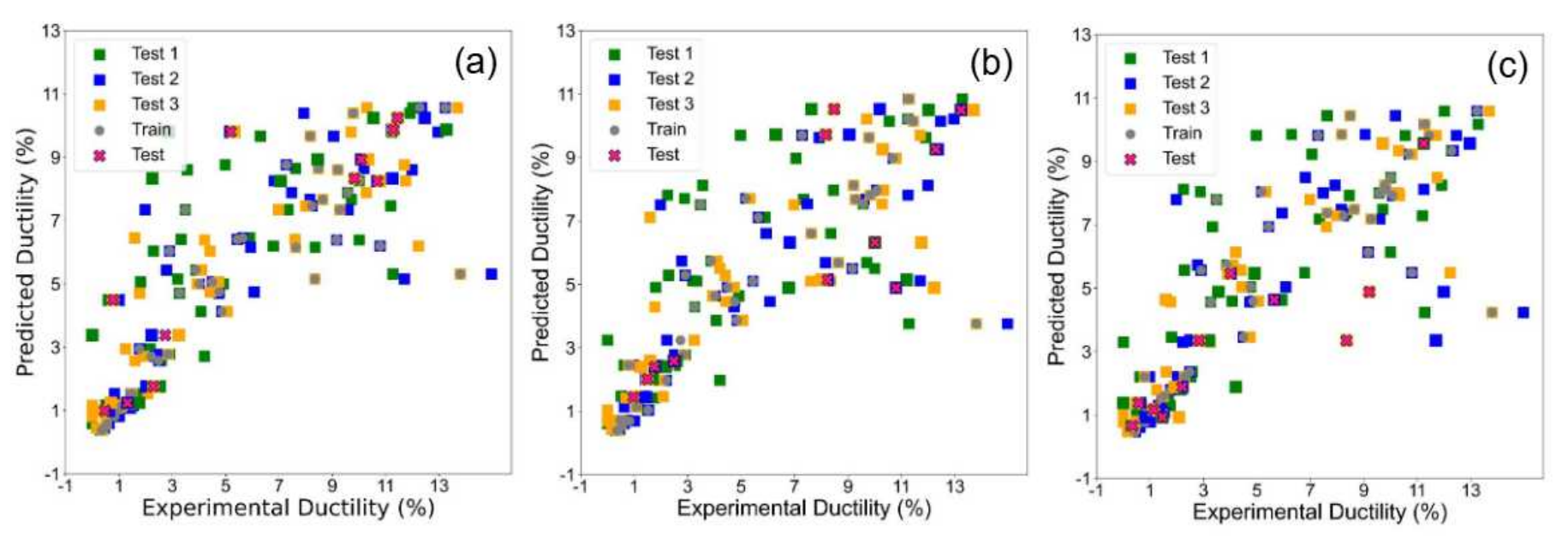}
    \vspace{-3mm}
    \caption{\textbf{Graphical representation of $5$-fold CV with $GP_{\varepsilon}$:} Each sample undergoes three repeated tensile tests. The squares represent the experimental ductility values obtained from each test case versus their predicted median. The pink ``X" symbols denote the model's predictions on test data, while the gray circles depict the model's performance on the training data. Each horizontal line in the plots highlights the variability of the test cases for each sample.}
    \label{fig: CV_ductility}
\end{figure}

We now perform a $5$-fold CV with $GP_{\varepsilon}$ which, similar to $GP_{\text{\YS}}$, also uses an FFNN as its mean function. Since the larger stochasticity of \D~compared to \YS~increases the risk of overfitting, we reduce the size of the FFNN to $2$ layers of $2$ neurons.
The results of the $5$-fold CV are shown in the bottom row of \Cref{tab: CV_YS} and demonstrate that $GP_{\varepsilon}$ is on average more accurate than $GP_{\varepsilon'}$. 
Comparing the estimated noise variance with the MSEs illustrates that ductility predictions \textit{expectedly} exhibit larger errors compared to other properties.
These variations are also illustrated in \Cref{fig: CV_ductility} where the colored square markers show the strain to failure of different test repetitions for the same sample versus their \D. Each group of three squares in a horizontal line corresponds to one process combination and highlights how ductility varies across three tensile tests. The pink ``X" markers and the grey circles indicate the predicted \D~for test and training data, respectively. 
Considering the high variability of \D, we deem these results satisfactory and proceed to build the final model for ductility on all tensile and cuboid samples. We denote this model by $GP_{\text{\D}}$ and test its generalizability in \Cref{subsec optimization}.

\subsection{Process Optimization and Design Maps} \label{subsec optimization}

We now use the trained GP models to identify the process parameters that optimize the combination of yield strength and ductility of a tensile coupon. 
In the literature, Bayesian optimization (BO) or metamodel-based searches are commonly employed for optimization but these methods are not suitable to our problem. BO finds the optimum of black-box functions by iteratively sampling the most informative points in the parameter space \cite{BO1,BO2,BO3, CMAME_SANAZ, JMD_Sanaz} while we aim to build and test only one sample. 
Metamodel-based methods leverage surrogates (GPs in our case) in an optimization package that are either gradient-based (e.g., Adam or L-BFGS) or heuristic (e.g., Genetic algorithm). We found these methods to provide poor performance in our case as their reported optima strongly depended on the initialization. We attribute this poor performance to the fact that we use four hierarchically linked GPs to predict tensile properties given the process parameters, and such a linkage produces a cascade of uncertainties that are not easy to quantify.



\begin{figure}[!b]
  \centering
    \includegraphics[width=\textwidth]{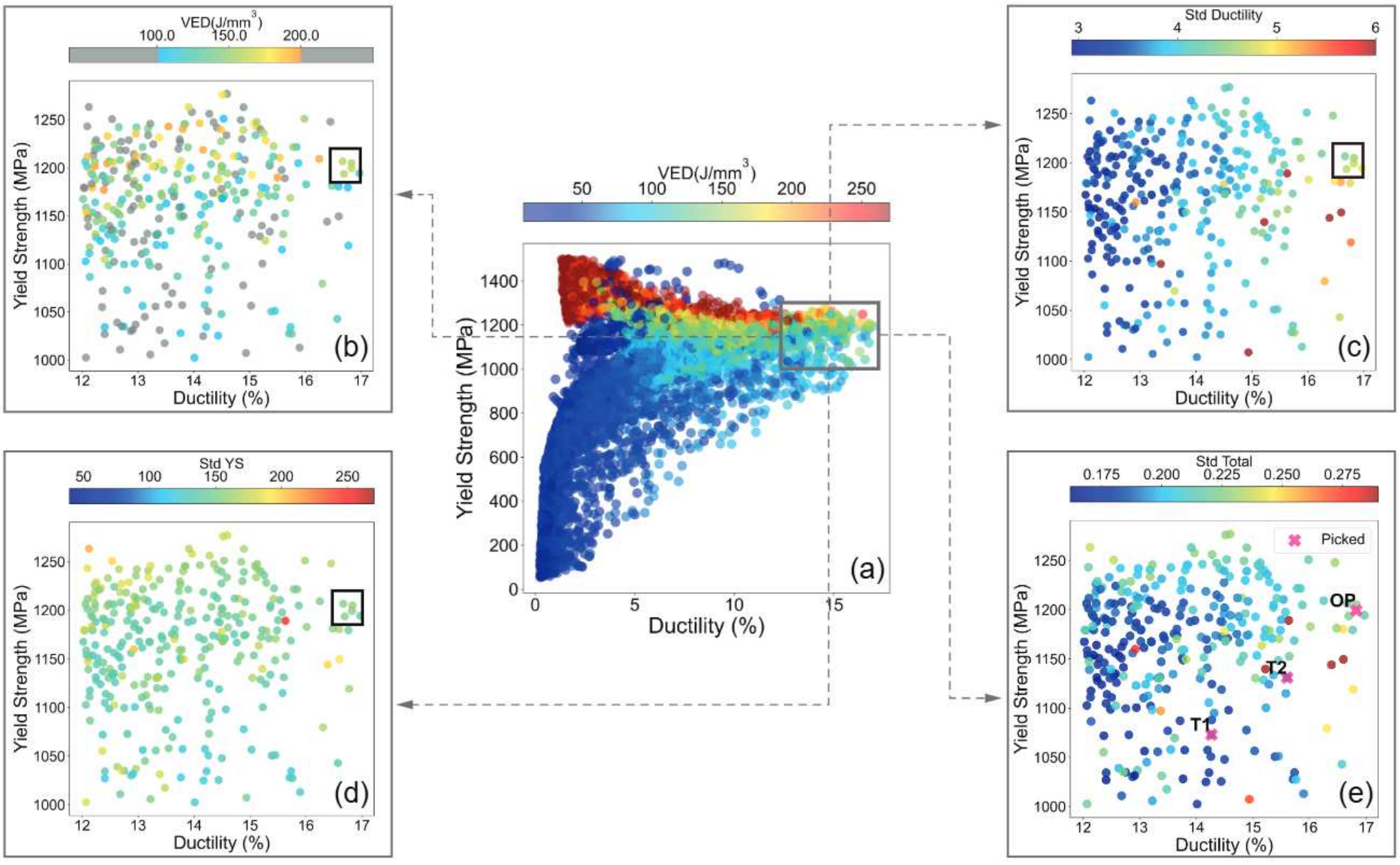}
    \vspace{-5mm}
    \caption{\textbf{Sampling for optimization:} \textbf{(a)} Predicted yield strength vs ductility for $10,000$ random sets of process parameters, color-coded based on VED. To identify optimal process setting, we focus on points with high yield strength ($\text{\hy} > 1000$) and ductility ($\text{\hd} > 12$). \textbf{(b)} Following the discussions in \Cref{subsec hardness porosity,subsec tensile}, process settings whose VED is outside the $100-200 J/mm^3$ range are colored gray. \textbf{(c)} and \textbf{(d)} color code the points based on the prediction uncertainties of yield strength and ductility, respectively. \textbf{(e)} illustrates the overall uncertainty in predicting yield strength and ductility. The three points marked as 'X' denote our selections for testing.}
    \label{fig: optim_map}
\end{figure}

We address the above challenges via the following simple and intuitive approach, see \Cref{fig: optim_map}. We first generate $10,000$ process parameter combinations and then use the trained GPs to predict the corresponding yield strength and ductility. The result is shown in \Cref{fig: optim_map}(a) where the points are color coded based on the VED. We then narrow the search space to include the points with $(1)$ VEDs between $100$ and $200$ J/mm$^3$ (based on the findings of \Cref{subsec hardness porosity,subsec tensile}), and $(2)$ high yet feasible \YS~($\text{\hy}>1000$ MPa) and \D~($\text{\hd}>12 \%$), see \Cref{fig: optim_map}(b). To consider the prediction uncertainties of the candidate points in \Cref{fig: optim_map}(b), we color code them based on the predicted uncertainties for the individual objectives (i.e., ductility and yield strength), see \Cref{fig: optim_map}(c) and (d). We finally identify process settings with small uncertainties, see the small black box in \Cref{fig: optim_map}(c) and (d), and randomly select one of them for the manufacturing process. The chosen setting is marked by ``OP" in \Cref{fig: optim_map}(e) where the points are color coded based on the uncertainty of the objective functions that considers both ductility and yield strength. We also select two random points outside of optimal window, marked as ``T1" and ``T2" in \Cref{fig: optim_map}(e), to assess the overall predictive power of our framework. For ``T2" we consider two scan rotations to evaluate our prediction based on the Sobol SIs that this process parameter insignificantly affects the mechanical properties. 

\begin{figure} [!b]
    \centering
    \includegraphics[width=1\linewidth]{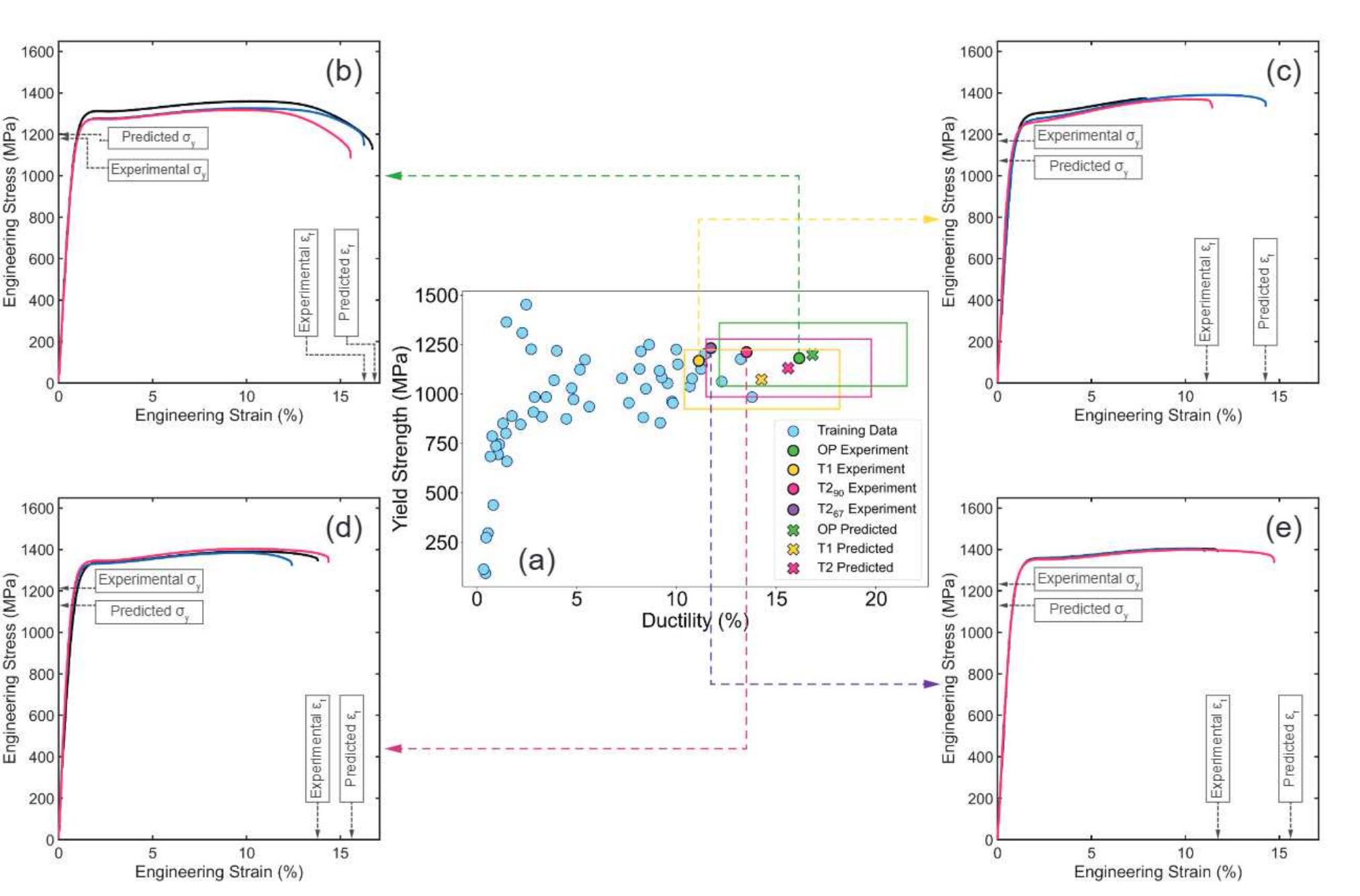}
    \caption{\textbf{The yield strength-ductility trade-off and the stress-strain curves of tested samples:} \textbf{(a)} YS-ductility trade-off for tensile and tested samples. The circles represent experimental samples, while the 'X' shapes indicate predicted samples. Rectangles display the $70\%$ confidence interval, with each rectangle colored to match its corresponding test sample. \textbf{(b)} Optimized condition, \textbf{(c)} T1, \textbf{(d)} T$2_{67}$, and \textbf{(e)} T$2_{90}$. Note: T$2_{90}$ and T$2_{67}$ are the same laser processing conditions with one printed with a scan rotation of $90$ and the other with $67$ degrees.}
    \label{fig strength-ductility tradeoff}
\end{figure}

The three selected process parameter sets are used to built and test tensile coupons, see \Cref{tab: optimized parameters}. The stress-strain curves of these samples are shown in \Cref{fig strength-ductility tradeoff} where we also provide the \YS-\D~scatter plot of the $54$ training samples to put the obtained results into perspective. 
It is observed that the experimental measurements (\yy~and \yd) match with model predictions (\hy~and \hd) quite well and the $70\%$ prediction intervals in all cases contain the experimental measurements. We also observe that changing the scan rotation from $90$ or $67$ degrees has an insignificant impact on the mechanical properties. This finding is consistent with the literature where $67$ is reported to show only slightly lower \YS, \UTS~and \D~for 316L SS \cite{Leicht2020EffectFusion}.

\begin{table}[!t]
\centering
\begin{tabular}{lllllllllll}
\hline
\multicolumn{1}{c}{{\begin{tabular}[c]{@{}c@{}}Sample \\ name\end{tabular}}} & \multicolumn{1}{c}{{$p$}} & \multicolumn{1}{c}{{$v$}} & \multicolumn{1}{c}{{$l$}} & \multicolumn{1}{c}{{$h$}} & \multicolumn{1}{c}{{$sr$}} & \multicolumn{1}{c}{{\begin{tabular}[c]{@{}c@{}}VED\\ (J/mm$^3$)\end{tabular}}} & \multicolumn{1}{c}{{\begin{tabular}[c]{@{}c@{}} \hy \\(MPa)\end{tabular}}} & \multicolumn{1}{c}{{\begin{tabular}[c]{@{}c@{}} \yy \\(MPa)\end{tabular}}} & \multicolumn{1}{c}{{\begin{tabular}[c]{@{}c@{}} \hd \end{tabular}}} & \multicolumn{1}{c}{\textbf{\begin{tabular}[c]{@{}c@{}} \yd \end{tabular}}} \\
\hline
{OP} & $81$ & $325$ & $20$ & $77$ & $90$ & $153.4$ & $1199$ & $1180.11$ & $16.82$ & $16.26$ \\
{T1} & $233$ & $1471$ & $20$ & $71$ & $90$ & $111.4$ & $1073$ & $1168.56$ & $14.26$ & $11.13$ \\
{T$2_{90}$} & $227$ & $1080$ & $20$ & $72$ & $90$ & $155.4$ & $1130$ & $1213.33$ & $15.60$ & $13.79$ \\
{T$2_{67}$} & $227$ & $1080$ & $20$ & $72$ & $67$ & $155.4$ & $1130$ & $1232.17$ & $15.60$ & $11.73$ \\
\hline
\end{tabular}
\caption{\textbf{Optimized processing conditions with corresponding predicted and experimentally measured tensile properties:} \yy~and \yd represent the experimental \YS~and \D, respectively, while \hy~and \hd~denote the corresponding predictions.}
\label{tab: optimized parameters}
\end{table}                          

As shown in \Cref{fig strength-ductility tradeoff}(a), it is observed that yield strength and ductility are not necessarily inversely correlated in our data, as is regularly the case for most conventionally processed structural materials. 
The strength-ductility trade-off is usually related to intrinsic strengthening mechanisms such as solid solution, precipitation, grain boundary, or second-phase strengthening which tend to reduce ductility. However, the large process parameter space that we explore can change this trend as defects, complex microstructures, and variations arise. Nonetheless, our results show that tuning the process parameters can increase both yield strength and ductility.

Compared to previous studies \cite{Yadollahi2017EffectsSteel, Auguste2018STUDYMELTING, Moyle2024CorrelationFusion, Salzbrenner2017High-throughputSteel, Pasebani2018EffectsMelting, Dong2023InfluenceSteel}, we have designed an LPBF processed \alloy~with improved tensile strength and ductility even though our tensile specimens were created vertically and the loading direction during the tensile test was the same as the build direction. Vertically-built specimens exhibit reduced ductility and tensile strength compared to those built horizontally \cite{Yadollahi2017EffectsSteel, Auguste2018STUDYMELTING, Moyle2024CorrelationFusion} since defects, which primarily form between layers, are perpendicular to the tensile loading direction in vertical specimens which, in turn, facilitates void growth \cite{Yadollahi2017EffectsSteel}. Additionally, the shorter time interval between melted layers and decreased cooling rates in vertical samples coarsen the grain sizes and lower retained austenite content \cite{Yadollahi2017EffectsSteel}, which reduces the strength. 
Also, our samples are tested with their surfaces in the as-printed condition without any machining or polishing as opposed to other studies where surface roughness is removed by grinding or polishing \cite{Li2022HomogenizationFusion, Dong2023InfluenceSteel}. Notwithstanding these strength/ductility reducing factors, we still achieved improved properties compared to the literature \cite{Yadollahi2017EffectsSteel, Auguste2018STUDYMELTING, Moyle2024CorrelationFusion, Salzbrenner2017High-throughputSteel, Pasebani2018EffectsMelting, Dong2023InfluenceSteel}.

\begin{figure}[!h]
  \centering
    \includegraphics[width=\textwidth]{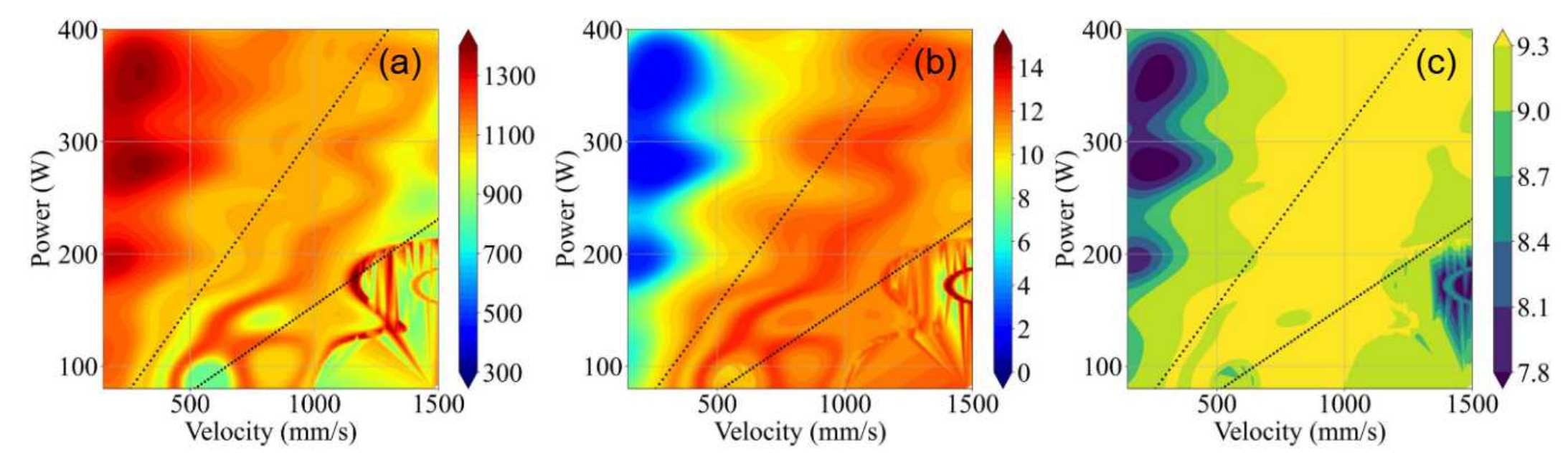}
    \vspace{-5mm}
    \caption{\textbf{Design maps for various combinations of $p$ and $v$:} The figures are color-coded based on \textbf{(a)} predicted \YS~(\hy), \textbf{(b)} predicted \D~(\hd), and \textbf{(c)} $log(\text{\hy})+log(\text{\hd})$ to show the optimization surface. Layer thickness and hatch spacing are fixed to the optimal values, $l=20$ $\mu$m and $h=77$ $\mu$m. The black dashed mark the VED values of $100$ and $200$ J/mm$^3$.}
\label{fig: design_maps}
\end{figure}

Finally, we use the trained GPs to plot design maps for the properties. \Cref{fig: design_maps} shows contour plots of \hy~and \hd~for different values of laser power and scan speed but fixed hatch spacing ($h=20$ $\mu$m) and layer thickness ($l=77$ $\mu$m) (see \Cref{fig: design_maps_si} for more contour plots). These plots are color-coded based on the property of interest, and the black dashed lines mark the VED values of  $100$ and $200$ J/mm$^3$.
The orange region in \Cref{fig: design_maps}(a) indicates the LPBF parameter window that leads to \hy~values around $1,100-1,200$ MPa. Similarly, the red region in \Cref{fig: design_maps}(b) is the LPBF parameter window that yields \hd~greater than $12\%$. To optimize the combination of \YS~and \D, the optimization surface (i.e., the logarithm of the products of \YS~and \D) is depicted in \Cref{fig: design_maps}(c), where the region in yellow marks the optimized process window.  


    \section{Conclusions} \label{Sec conclusion}
We develop a process optimization framework that integrates HT experiments and hierarchical ML to identify the processing parameters that optimize the mechanical properties of LPBF-built parts. Our approach reduces the reliance on expensive tensile tests and streamlines process optimization by systematically integrating computational techniques (statistical sensitivity analysis, feature augmentation, data fusion, and emulation) with experiments that include large data from cuboid samples and small data from tensile coupons. While the proposed approach is material-agnostic and fully generic, in this paper we apply it to \alloy~which is a technologically important material.

We demonstrate that conventional process optimization approaches that solely rely on hand-crafted features, such as VED, fail to fully capture the effects of process parameters on the material properties and limit the search to small parameter spaces. Our framework addresses these gaps because it directly and automatically predicts mechanical properties as functions of all processing parameters including laser power, laser scan speed, hatch spacing, powder layer thickness, and scan rotation. 
Additionally, by founding these predictions on larger ``easy to collect'' and smaller ``labor-intensive'' data sets, our approach is much less dependent on expensive fabrication and characterization procedures, enabling the exploration of a large set of process parameters.


We demonstrate that our framework is very efficient at learning the complex relationships between multiple processing parameters (varying in very wide ranges) and tensile mechanical properties. Following model training and validation, we identify near-optimal processing conditions. By printing tensile specimens with these conditions, we $(1)$ demonstrate that our GPs predict both \YS~and \D quite accurately, and $(2)$ can build vertically printed tensile specimens with extreme combinations of strength and ductility compared to typical literature data. 

Unsurprisingly, we observe that the mechanical properties of LPBF processed \alloy~including hardness, \YS, \UTS, and especially \D~, strongly depend on processing parameters.
While the present work focused on the development of a novel approach for learning these complex correlations and optimizing processing parameters, ML models cannot provide physics-based explanations. Future work will carefully investigate the subtle microstructural differences among some of the samples fabricated in this study, to reach a full mechanistic understanding of the processing-structure-properties relationships for this complex material system.



\section*{Acknowledgments}
This research was supported by funds from the UC National Laboratory Fees Research Program of the University of California, Grant No. L22CR4520. 
M.A. and L.V. also acknowledge financial support by the Oﬃce of Naval Research (Program Manager: J. Wolk, Grant No. N00014-21-1-2570).
Z.Z.F. and R.B. also acknowledge the support from the National Science Foundation (Award No. 2238038).
    \begin{appendices}

\setcounter{equation}{0}
\renewcommand{\theequation}{\thesection-\arabic{equation}}

\setcounter{figure}{0}
\renewcommand\thefigure{\thesection\arabic{figure}}
\setcounter{table}{0}
\renewcommand\thetable{\thesection\arabic{table}}

\section{Porosity Measurement} \label{sec: porosity_measurement}
\begin{figure}[!t]
  \centering
    \includegraphics[width=0.6\textwidth]{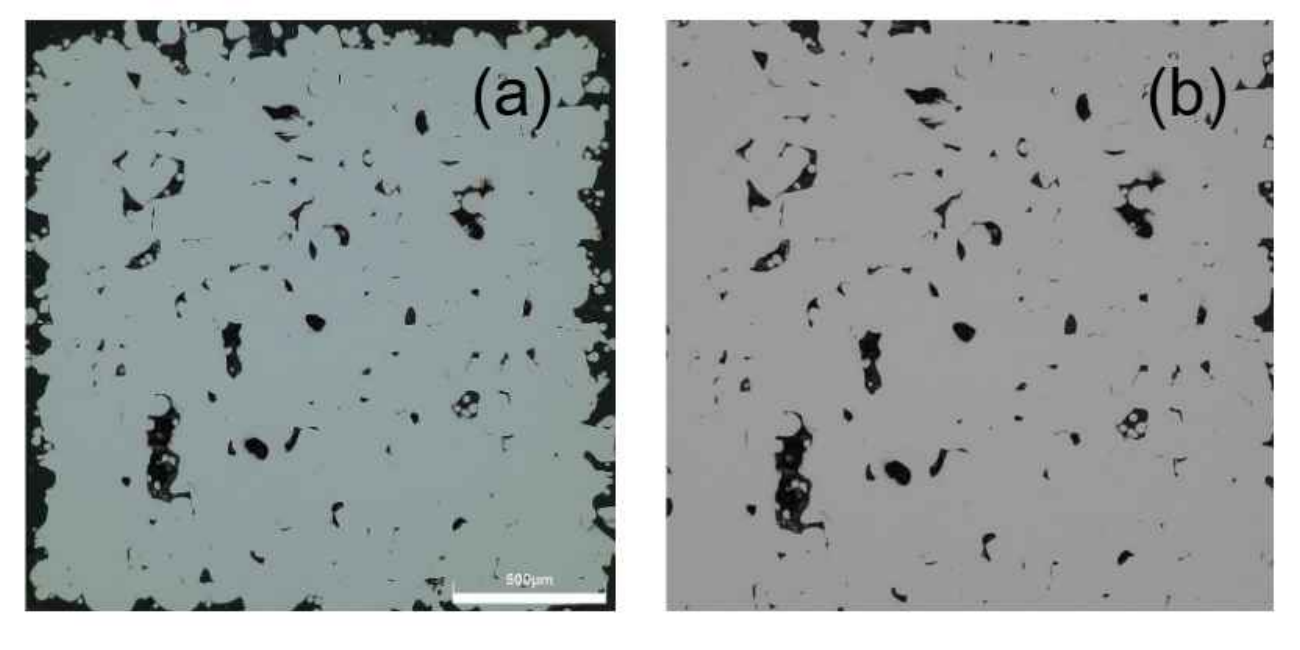}
    \vspace{-3mm}
    \caption{\textbf{OM images of cuboid sample $57$:} \textbf{(a)} shows the original OM image for this sample while \textbf{(b)} illustrates the cropped and blurred image that we use through our analysis.}
    \label{fig: micros}
\end{figure}

In this section, we provide details of the porosity measurement approach used to extract the porosity content of each sample based on the OM images obtained from the as-polished surfaces. To enhance image quality and minimize noise interference, we employ preprocessing techniques such as cropping and Gaussian blurring.
Firstly, for cropping, we remove $50$ pixels from the top, right, and left edges, and $80$ pixels from the bottom edge of each image. This adjustment is necessary due to the larger frame present in the bottom part of the images.
Additionally, we apply Gaussian blurring to further reduce noise. Gaussian blurring involves averaging pixel values using a Gaussian kernel, which assigns higher weights to pixels closer to the center. This method effectively smooths the image and improves clarity for subsequent analysis. We utilize a Gaussian kernel with a size of $5 \times 5$ to define the pixel neighborhood for blurring. \Cref{fig: micros} shows these process for a random cuboid sample (sample $57$) where \Cref{fig: micros}(a) is the original OM image and \Cref{fig: micros}(b) shows the preprocessed image that we use for our analysis. 

\begin{figure}[!b]
  \centering
    \includegraphics[width=0.8\textwidth]{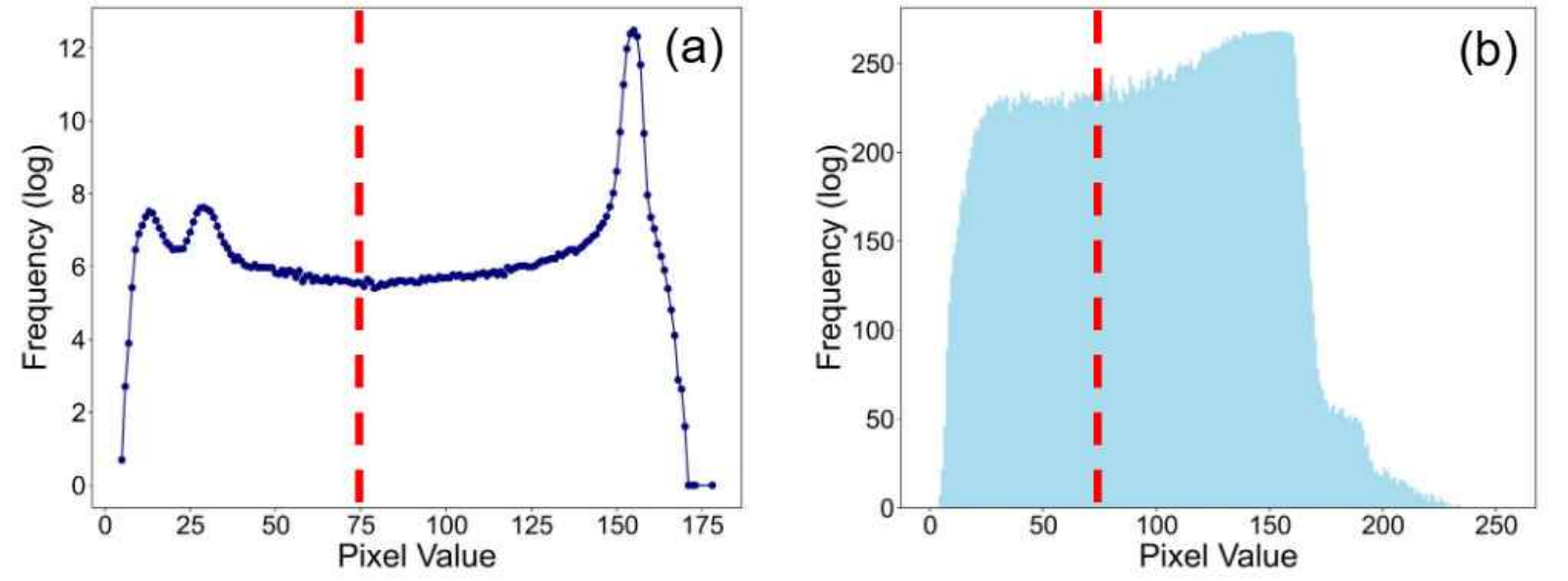}
    \vspace{-3mm}
    \caption{\textbf{Pixel distribution of OM images:} \textbf{(a)} shows the pixel distribution for sample $57$ while \textbf{(b)} illustrates the distribution for all samples. }
    \label{fig: pixel_dist}
\end{figure}

After preprocessing, we analyze the distribution of pixel values to determine an appropriate threshold for distinguishing between pores and solid materials. Figure \ref{fig: pixel_dist}(a) illustrates this distribution for cuboid sample $57$, with frequencies reported on a logarithmic scale for clarity. The distribution reveals two peaks: one in the range of $20-30$, indicative of frequent pixel values for pores, and another around $150-160$, representing solid material.
Figure \ref{fig: pixel_dist}(b) displays the aggregate distribution across all cuboid samples with the same trend as sample $57$. Based on these observations, we select a threshold value of $75$ (a middle value) to differentiate between pores and solid material. Finally, porosity is computed as the ratio of pore area to the total image area. All the measured porosities are provided in \Cref{tab: parameter sets}.

\section{Background on Gaussian Processes (GPs)} \label{sec: back_GP_appendix}
In emulation using GPs, it is assumed that the training data follows a multivariate normal distribution characterized by parametric mean and covariance functions. Predictions are then made using closed-form conditional distribution formulas.

Assume we are given a training dataset $\braces{\xb^{(i)}, y^{(i)}}_{i=1}^n$, where $\xb = [x_1, ..., x_{dx}]^T \in \xsspace \subset \rsspace^{dx}$ and $y^{(i)} = y(\xb^{(i)}) \in \rsspace$ represent the inputs and outputs, respectively. Let $\yb = [y^{(1)}, \cdots, y^{(n)}]^T$ and $\Xb$ be the matrix whose $i^{th}$-th row is $(\xb^{(i)})^T$. Our objective is to predict $y(\xb^*)$ at an arbitrary point $\xb^* \in \xsspace$.
Following this setup, we assume $\yb = [y^{(1)}, \cdots, y^{(n)}]^T$ is a realization of a GP characterized by the following parametric mean and covariance functions:

\begin{subequations} 
    \begin{equation} 
        \begin{split}
            \mathbb{E}[y(\xb)] = m(\xb; \betab),
        \end{split}
        \label{eq: gp-mean}
    \end{equation}
    \begin{equation} 
        \begin{split}
            \text{cov}\left(y(\xb), y(\xb')\right) = c(\xb, \xb'; \sigma^2, \thetab) = \sigma^2 r(\xb, \xb'; \thetab)
        \end{split}
        \label{eq: gp-cov}
    \end{equation}
    \label{eq: gp-mean and cov}
\end{subequations} 
\noindent where $\mathbb{E}[\cdot]$ indicates expectation, and $\betab$ and $\thetab$ are the parameters of the mean and covariance functions, respectively. 
The mean function in GP modeling, as seen in \Cref{eq: gp-mean}, can take various forms, from simple polynomials to intricate structures like feed-forward neural networks (FFNN). However, many GP applications often use a constant mean function $m(\boldsymbol{x}; \boldsymbol{\beta}) = \beta$, suggesting that the predictive power of the GP mainly depends on its kernel function. In \Cref{eq: gp-cov}, $\sigma^2$ stands for the process variance, and $r(\cdot, \cdot)$ is the correlation function with parameters $\boldsymbol{\theta}$. Popular choices for $r(\cdot, \cdot)$ include the Gaussian, power exponential, and Matérn correlation functions. In our specific approach, we utilize the Gaussian kernel.
\begin{equation} 
    r(\boldsymbol{x}, \boldsymbol{x}^{\prime}; \omegab)=\exp \left\{-\sum_{i=1}^{dx} 10^{\omega_{i}}(x_{i}-x_{i}^{\prime})^{2}\right\}
    \label{eq: gaussian-kernel}
\end{equation}
\noindent where $\omega_i \in \rsspace$ are the scale parameters.
However, the abovementioned formulations do not inherently support data fusion. Motivated by \cite{oune2021latent,gp+}, kernel-based approaches can be used to extend GPs to handle data fusion. 
This method introduces new kernels with customized parametric functions to enable direct probabilistic learning from multi-source data and handling qualitative features.

To explain the kernel-based approach, consider an emulation scenario where the input space includes two qualitative features: $ t_1=\{Cuboid, Tensile\} $ and $ t_2=\{174,316,304\}$, with $l_1 = 2$ and $l_2 =3$ levels, respectively. GPs cannot directly handle $\tb = [t_1, t_2]^T$ because typical kernels require a distance metric for each feature, which categorical variables inherently lack.
To overcome this limitation, categorical variables must be transformed into a quantitative representation ($\pib_t$) using a user-defined function ($f_{\pi}(\tb)$); thus,  $\pib_t = f_{\pi}(\tb)$. These quantitative representations can be generated using methods such as grouped one-hot encoding, separate one-hot encoding, or random encoding. In this paper, we employ grouped one-hot encoding.
These representations are typically high-dimensional. To reduce the dimensionality while capturing the effects of $\tb$ on the response, $\pib_t$ is processed through a parametric embedding function $f_h(\pib_t; \thetab_h)$ to generate $\hb$, a $dh$-dimensional latent representation of $\tb$, where $d\pi \gg dh$. The embedding functions can be either parametric matrices or FFNNs. In our approach, we utilize parametric matrices to generate $\hb$ as follows:

\begin{equation}
    \hb= \pib_t \times \boldsymbol{A}
\end{equation}

\noindent where $\boldsymbol{A}$ is a $\sum_{i=1}^{dt} l_i \times dh$ parametric mapping matrix that
maps $\pib_t$ (grouped one hot-encoded prior) to $h$.
Since $\hb = f_h(f_\pi(\tb); \thetab_h)$ are quantitative, they can be easily used to develop new kernels. 
Now, we can rewrite the kernel in \Cref{eq: gaussian-kernel} as:

\begin{equation} 
    \begin{split}
        r\left(\ub, \ub^{\prime}; \omegab, \thetab_h\right) = 
        \exp \left\{-\sum_{i=1}^{dx} 10^{\omega_i}(x_i-x_i^{\prime})^2 -
        \sum_{i=1}^{dh}(h_i - h^{\prime}_i)^2 \right \} 
    \end{split}
    \label{eq: gaussian-kernel-GP_Plus}
\end{equation}

\noindent where $\ub = \left[ \begin{array}{l} \xb \\ \tb \end{array} \right]$. The new parameters ( $\thetab_h$) will be estimated jointly with the other parameters of the GP.

Having defined the new kernel, all the hyperparameters ($\betab$, $\sigma^2$, $\thetab$) will be estimated via the training data. To find these estimates, we utilize maximum a posteriori (MAP) which estimates the hyperparameters such that they maximize the posterior of the $n$ training data being generated by $y(\boldsymbol{x})$, that is:

\begin{equation} 
    \begin{split}
        [\widehat{\betab}, \widehat{\sigma^2}, \widehat{\thetab}] = 
        \underset{\betab, \sigma^2, \thetab}{\operatorname{argmax}}\left|2 \pi \Cb\right|^{-\frac{1}{2}} \times 
        \exp \left\{\frac{-1}{2}(\yb-\mb)^T \Cb^{-1}(\yb-\mb)\right\}
        \times p(\betab, \sigma^2, \thetab)
    \end{split}
    \label{eq: map-max}
\end{equation}
or equivalently:
\begin{equation} 
    \begin{split}
        [\widehat{\betab}, \widehat{\sigma}^2, \widehat{\thetab}] = \underset{\betab, \sigma^2, \thetab}{\operatorname{argmin}} \hspace{2mm} {L_{MAP}}=
        \underset{\betab, \sigma^2, \thetab}{\operatorname{argmin}} \hspace{2mm} \frac{1}{2} \log (|\Cb|)+\frac{1}{2}(\yb-\mb)^T \Cb^{-1}(\yb-\mb)- \log\left(p(\betab, \sigma^2, \thetab)\right)
    \end{split}
    \label{eq: map-gp}
\end{equation}
\noindent where $|\cdot|$ denotes the determinant operator, $\Cb_{nn}:=c(\Xb, \Xb; \sigma^2, \thetab)$ is the covariance matrix whose $(i, j)^{t h}$ element is $C_{i j} = c(\xb^{(i)}, \xb^{(j)}; \sigma^2, \thetab) = \sigma^2r(\xb^{(i)}, \xb^{(j)}; \thetab)$, $\mb$ is an $n \times 1$ vector whose $i^{th}$ element is $m_i=m(\xb^{(i)}; \betab)$, and  $\log (\cdot)$ denotes the natural logarithm.
We can now efficiently estimate all the model parameters by minimizing \Cref{eq: map-gp} using a gradient-based optimization algorithm. Subsequently, we utilize the conditional distribution formulas to obtain the mean and variance of the response distribution at an arbitrary point $\xb^*$:
\begin{subequations} 
    \begin{equation} 
        \E[y(\xb^*)] =
        m(\xb^*; \widehat{\betab}) + c(\xb^*, \Xb; \widehat{\thetab}, \widehat{\sigma}^2) \boldsymbol{C}^{-1}(\yb-\mb)
        \label{eq: gp-mean-scalar}
    \end{equation}
    \begin{equation} 
        \text{cov}(y(\xb^*), y(\xb^*)) =
        c(\xb^*,\xb^*; \widehat{\thetab}, \widehat{\sigma}^2) - c(\xb^*, \Xb; \widehat{\thetab}, \widehat{\sigma}^2) \Cb^{-1} c(\Xb, \xb^*; \widehat{\thetab}, \widehat{\sigma}^2))
        \label{eq: gp-var-scalar}
    \end{equation}
    \label{eq: gp-mean-var-scalar}
\end{subequations}    
where $c(\xb^*, \Xb; \widehat{\thetab},\widehat{\sigma}^2)$ is a $1 \times n$ row vector with entries $c_i = c(\xb^*, \xb^{(i)}; \widehat{\thetab},\widehat{\sigma}^2)$ and its transpose is $c(\Xb, \xb^*; \widehat{\thetab},\widehat{\sigma}^2)$.
These formulations build interpolating GPs. 

To handle datasets with noisy observations, the nugget or jitter parameter, denoted by $\delta$ \cite{RN1917, RN332, RN1908}, is used. Accordingly, $\Cb$ is replaced by $\Cb_{\delta}= \Cb + \delta \Ib_{nn}$, where $\Ib_{n×n}$ is the $n \times n$ identity matrix. With this adjustment, the stationary noise variance estimated by the GP is $\widehat\delta$). Following this modification, \Cref{eq: gp-mean-var-scalar} should be updated to:

\begin{subequations} 
    \begin{equation} 
        \E[y(\Xb^*)] = 
        m(\Xb^*; \widehat{\betab}) + c(\Xb^*, \Xb; \widehat{\thetab}, \widehat{\sigma}^2) \Cb^{-1}_{\delta}(\yb-\mb)
        \label{eq: gp-mean-vec-noise}
    \end{equation}
    \begin{equation} 
        \text{cov} \parens{y\parens{\Xb^*}, y(\Xb^*)} =
        c(\Xb^*, \Xb^*; \widehat{\thetab}, \widehat{\sigma}^2) -
        c(\Xb^*, \Xb; \widehat{\thetab}, \widehat{\sigma}^2) \Cb^{-1}_{\delta} 
        c(\Xb, \Xb^*; \widehat{\thetab}, \widehat{\sigma}^2) + \widehat\delta\Ib.
        \label{eq: gp-var-vec-noise}
    \end{equation}
    \label{eq: gp-mean-var-vec-noise}
\end{subequations}    

The above kernel reformulations not only enable GPs to operate in feature spaces with categorical variables, but they also allow GPs to directly fuse multiple datasets from various sources. Suppose we have $ds$ data sources and aim to emulate all of them. To achieve this, we first augment the input space with an additional categorical variable $s=\{'1', \cdots, 'ds'\}$, where the $j^{th}$ element corresponds to data source $j$ for $j = 1, \cdots, ds$. With this augmentation, the $ds$ datasets are concatenated as follows:
\begin{equation} 
    \begin{split}
        \boldsymbol{U}=\left[\begin{array}{cc}
        \boldsymbol{U}_1 & '\mathbf{1}'_{n_{1} \times 1} \\
        \boldsymbol{U}_2 & '\mathbf{2}'_{n_{2} \times 1} \\
        \vdots & \vdots \\
        \boldsymbol{U}_{ds} & '\mathbf{ds}'_{n_{ds} \times 1}
        \end{array}\right] 
        \text { and }
        \boldsymbol{y}=\left[\begin{array}{c}
        \boldsymbol{y}_1 \\
        \boldsymbol{y}_2 \\
        \vdots \\
        \boldsymbol{y}_{ds}
        \end{array}\right]          
    \end{split}
    \label{eq: GP_Plus-fidelity-append}
\end{equation}
\noindent where the subscripts $1, 2, ..., ds$ correspond to the data sources, $n_j$ is the number of samples obtained from source $j$, $\boldsymbol{U}_j$ and $\boldsymbol{y}_j$ are, respectively, the $n_j \times (dx + dt)$ feature matrix and the ${n_j \times 1}$ vector of responses obtained from $r(j)$, and $'\boldsymbol{j}'$ is a categorical vector of size ${n_j \times 1}$ whose elements are all set to $'j'$. Then, the GP model can be trained on $\{ \boldsymbol{U}, \boldsymbol{y} \}$ using the kernel-based method described above to effectively integrate and fuse data.

Modifying the mean function presented in \Cref{eq: gp-mean} has a significant impact on the model's performance, especially in tasks such as extrapolation and data fusion. To enhance emulation, existing methods often use polynomials, analytic functions (e.g., $\sin(\cdot), \log(\cdot), \cdots$, etc.), or feedforward neural networks (FFNNs) for designing $m(\xb; \betab)$. Extending these approaches to also include categorical variables in the mean function ($m(\xb,\tb; \betab)$) enables two distinct learning strategies: $(1)$ employing a global function shared across all data sources, and $(2)$ using mixed basis functions where a unique mean function is learned for each data source $r$. The latter strategy presents a significant advantage for our problem. By estimating unique mean functions, the model can more accurately capture the specific behaviors of each data source, while the joint estimation of the functions' parameters allows the model to learn the interdependencies among the data sources.

All the emulation strategies mentioned in this section are available in an open-source Python package \gp, which we utilize for implementation.

\section{Sensitivity Analysis} \label{sec Sobole}
Evaluating the sensitivity of the output to the input features is essential as it can aid in feature selection.
Sobol sensitivity analysis is a global variance-based method used for quantifying each input’s main and total contribution to the output variance \cite{saltelli2010variance}. While main-order Sobol indices (SIs) reveal the individual contributions of input variables, total-order indices capture both the individual and interaction effects of inputs on the output.

\begin{table}[!b]
    \centering
    \begin{tabular}{c|cccc}
    \setlength{\tabcolsep}{1pt}
        & \multicolumn{1}{c}{} & Features &\\ \hline
        \multicolumn{1}{c|}{Metric} &\multicolumn{1}{c}{$x_1$}& \multicolumn{1}{c}{$x_2$} & \multicolumn{1}{c}{$x_3$} & \multicolumn{1}{c}{$x_4$} \\ \hline
        \multicolumn{1}{c|}{Main SI}  & $0.2315$       & $0.2384$      & $0.2355$ & $0.0001$ \\
        \multicolumn{1}{c|}{Total SI}  & $0.5266$       & $0.5333$      & $0.2358$ & $0.0006$ \\
        \hline    
    \end{tabular}
    \caption{\textbf{Sensitivity analysis:} Sensitivity analysis of $y(\boldsymbol{x}) = x_1^2 + x_2^2 + x_1 x_2 + x_3^2 + 10^{-3} \times x_4^2$ with $-1 < x_i < 1$ using Sobol indices.}
    \label{table: SA_borehole_test}
\end{table}

To gain deeper insights into these indices, we examine a $4$-dimensional case where $y(\boldsymbol{x}) = x_1^2 + x_2^2 + x_1 x_2 + x_3^2 + 10^{-3} \times x_4^2$, with each variable constrained to $-1 < x_i < 1$. The results in \Cref{table: SA_borehole_test} showcase the computed total and main SIs for this scenario. Notably, according to the table, $x_4$ exhibits the lowest total and main SIs, indicating its minimal impact on the output. This limited effect can be attributed to its very small coefficient ($10^{-3}$), which significantly mitigates $x_4$'s effect on the output variance.

The main SIs for $x_1$, $x_2$, and $x_3$ are close, reflecting that these variables have similar individual contributions to the output variance. This is consistent with their roles in the function, where $x_1^2$, $x_2^2$, and $x_3^2$ are equally weighted terms affecting the output directly.
However, despite their comparable individual effects, $x_1$ and $x_2$ exhibit significantly larger total effects ($0.5266$ and $0.5333$, respectively) compared to $x_3$. This discrepancy is due to the interaction term $x_1 \times x_2$  in the function, which increases the total contribution of both $x_1$ and $x_2$ beyond their individual effects.

In this paper, we employ Sobol analysis provided by \gp. Traditionally, Sobol sensitivity analysis is applied to quantitative features, but in our case, we also need to calculate it for categorical features due to the mixed nature of our process parameters. \gp~extends this analysis to include categorical features by sampling random quantitative values and associating them with the distinct levels of categorical variables. This method allows for a comprehensive evaluation of output sensitivity to both quantitative and categorical inputs, thereby enhancing the capability for more thorough feature selection.

\end{appendices}

    \bibliographystyle{unsrt} 
    \bibliography{01_Ref.bib}     
    \pagebreak 
    \section{Supplementary Information}
\beginsupplement

\begin{table}[p]
    \centering
    \includegraphics[width=\textwidth]{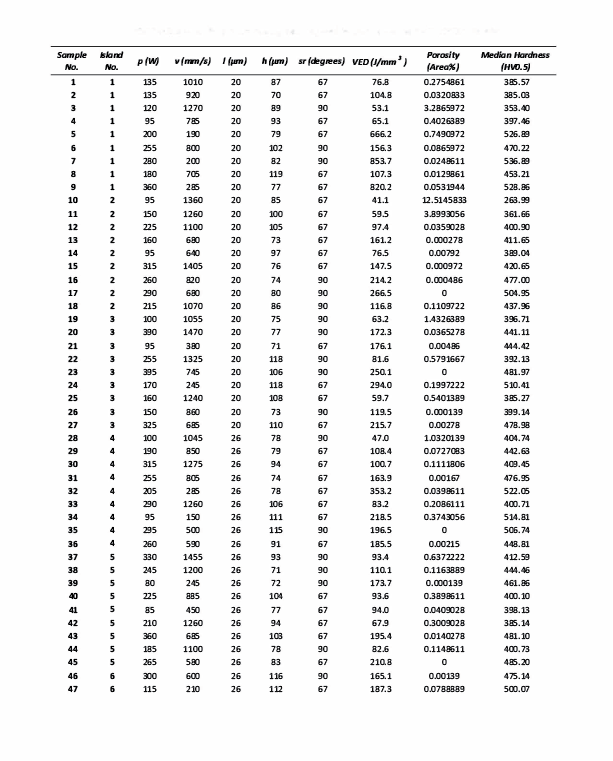}
    \caption{\textbf{Processing parameters, porosity and hardness values for the 270 cuboids.}}
    \label{tab: processing parameters}
\end{table}

\begin{table}[p]
    \ContinuedFloat
    \centering
    \includegraphics[width=\textwidth]{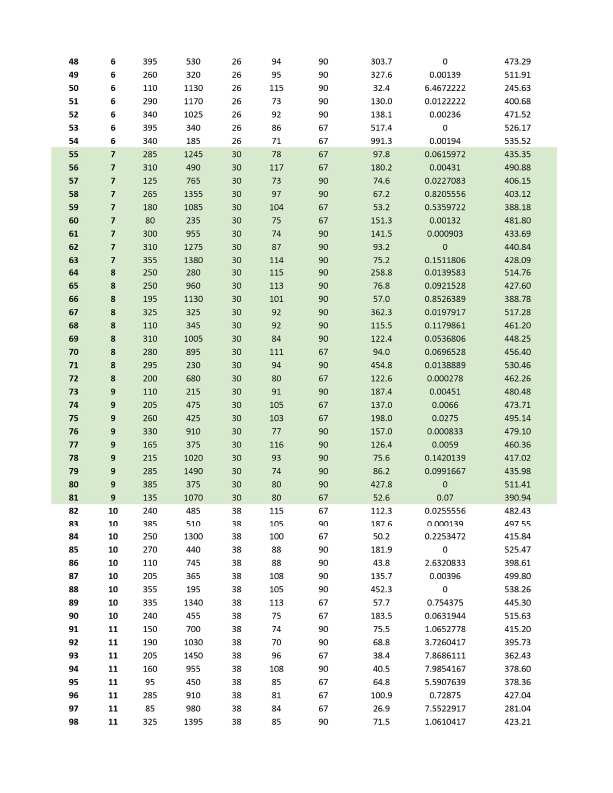}
    \caption[]{}
\end{table}

\begin{table}[p]
    \ContinuedFloat
    \centering
    \includegraphics[width=\textwidth]{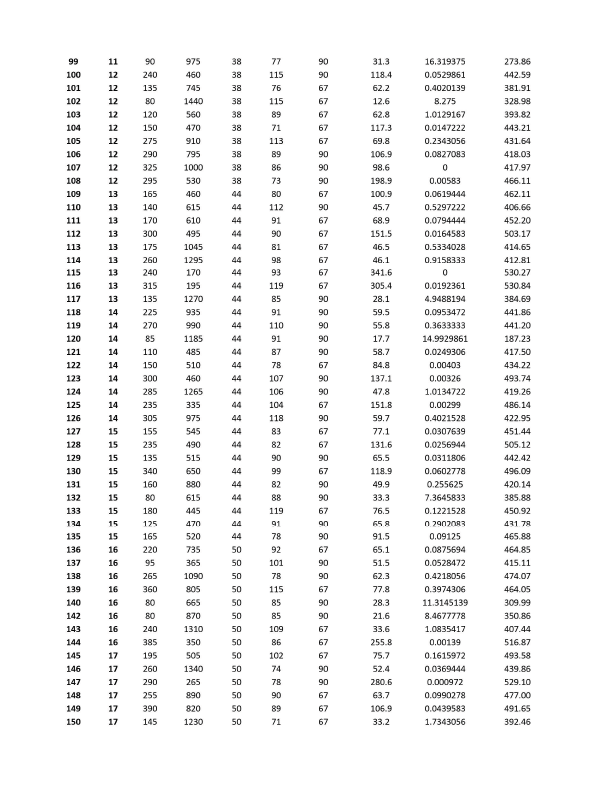}
    \caption[]{}
\end{table}

\begin{table}[p]
    \ContinuedFloat
    \centering
    \includegraphics[width=\textwidth]{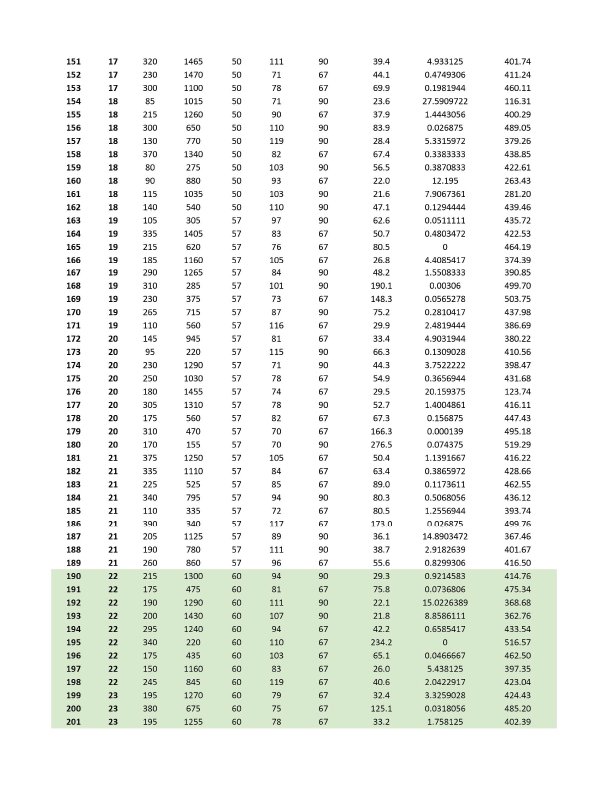}
    \caption[]{}
\end{table}

\begin{table}[p]
    \ContinuedFloat
    \centering
    \includegraphics[width=\textwidth]{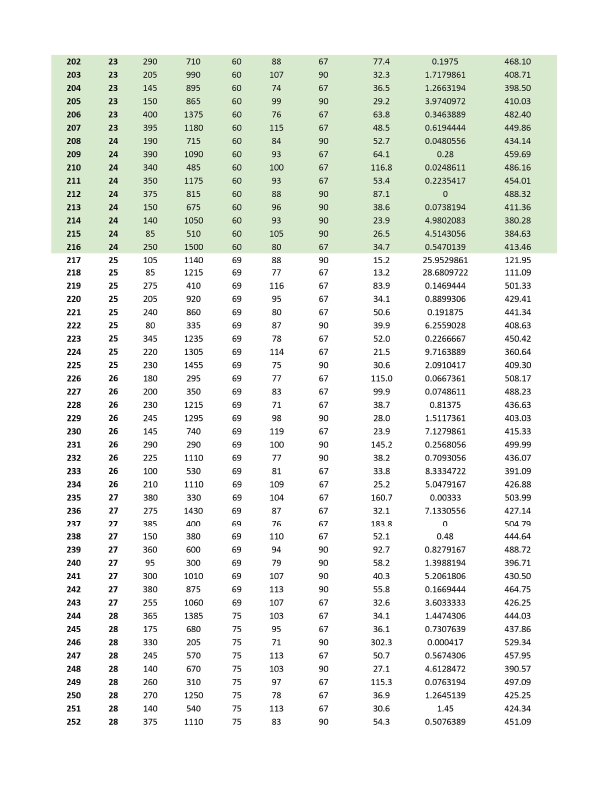}
    \caption[]{}
\end{table}

\begin{table}[p] \label{tab: parameter sets}
    \ContinuedFloat
    \centering
    \includegraphics[width=\textwidth]{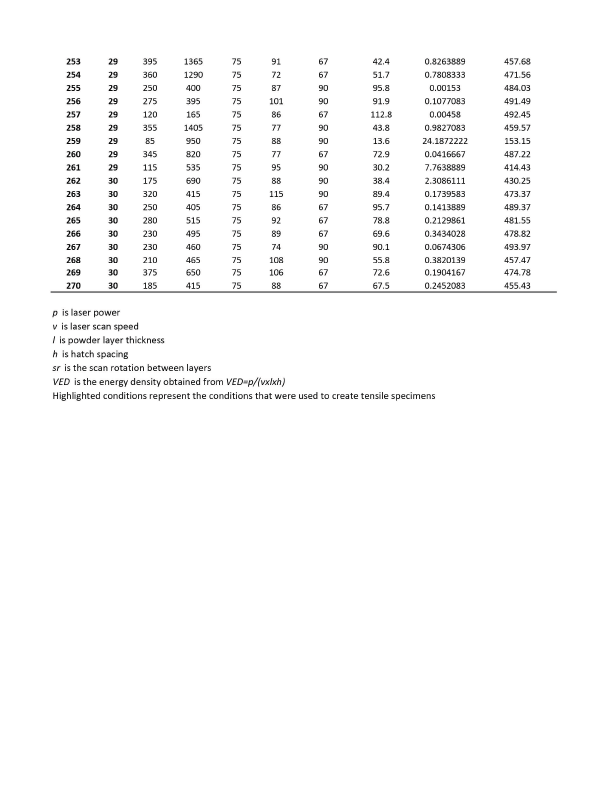}
    \caption[]{}
\end{table}



\begin{figure}[p]
    \centering
    \includegraphics[width=0.95\textwidth]{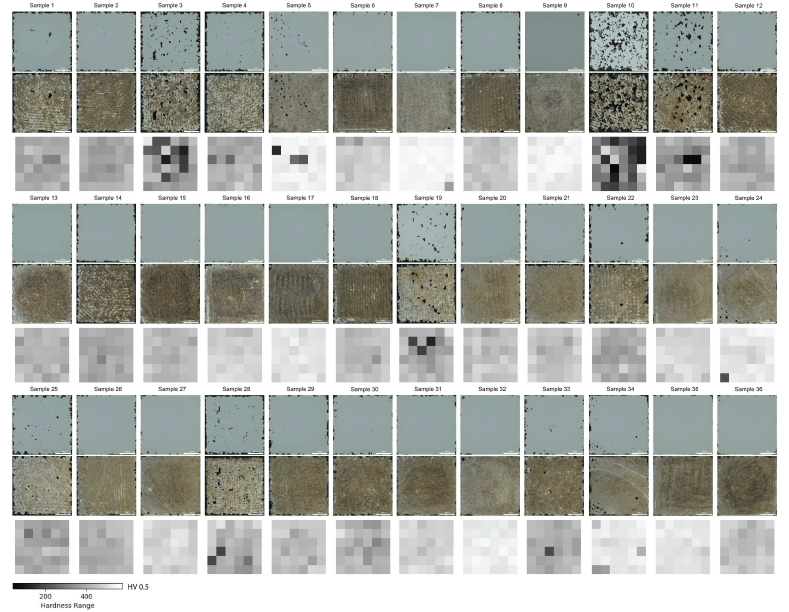}
    \vfill
    \includegraphics[width=0.95\textwidth]{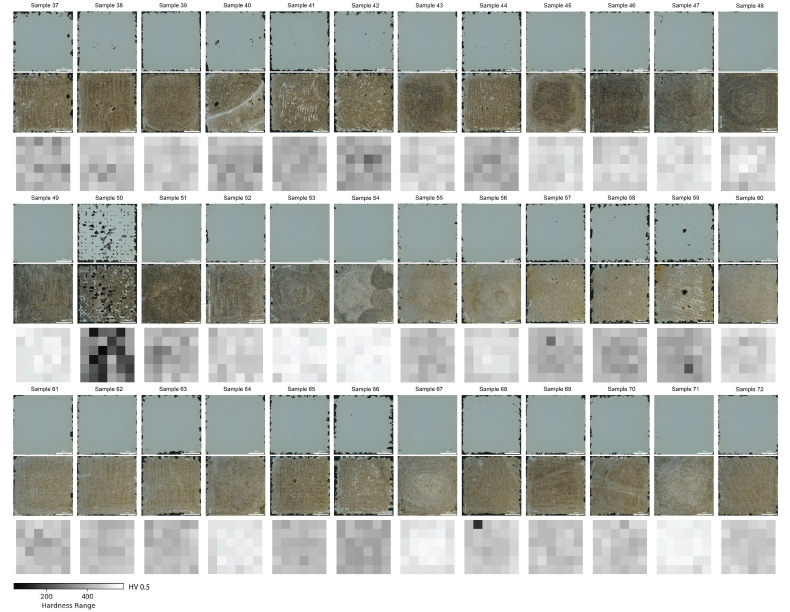}
    \caption[]{}
\end{figure}

\begin{figure}[p]
    \ContinuedFloat
    \centering
    \includegraphics[width=0.95\textwidth]{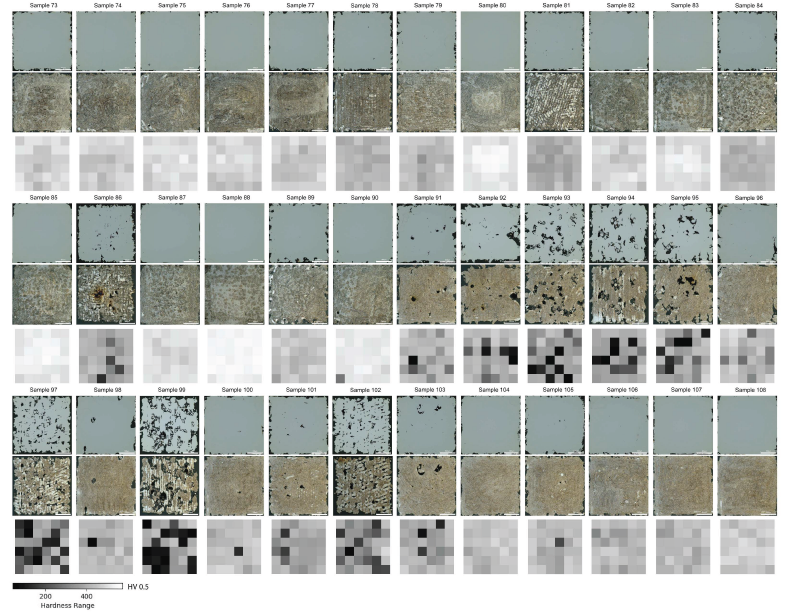}
    \vfill
    \includegraphics[width=0.95\textwidth]{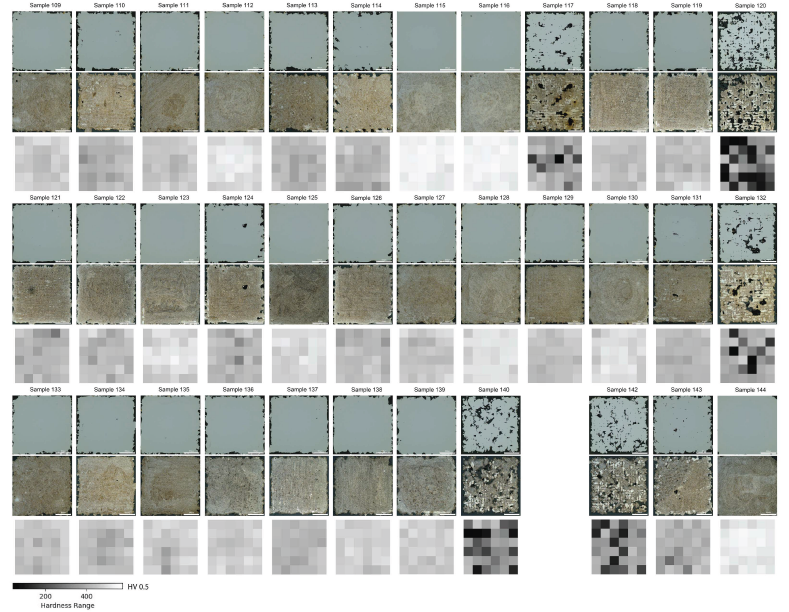}
    \caption[]{}
\end{figure}

\begin{figure}[p]
    \ContinuedFloat
    \centering
    \includegraphics[width=0.95\textwidth]{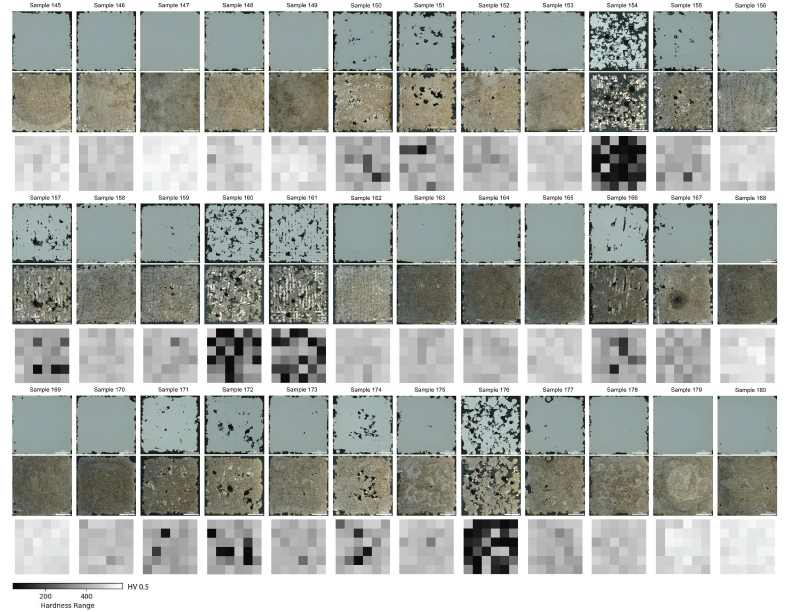}
    \vfill
    \includegraphics[width=0.95\textwidth]{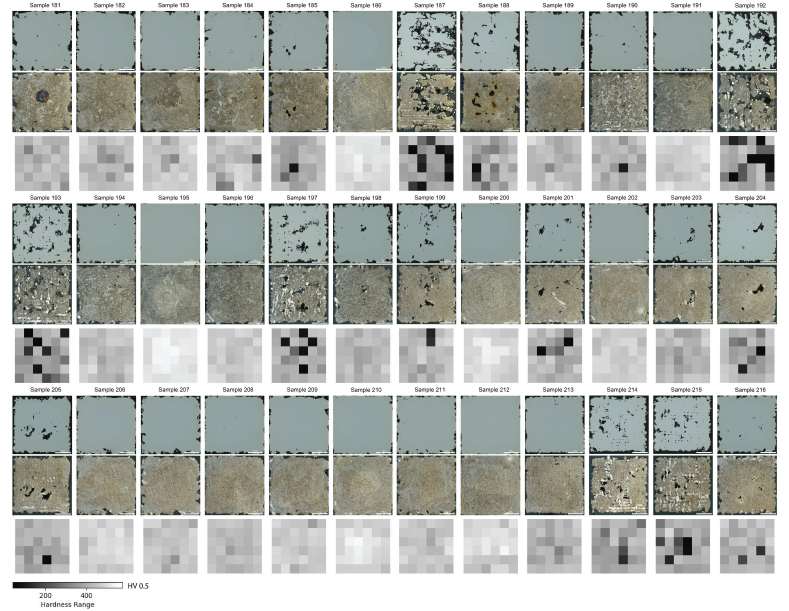}
    \caption[]{}
\end{figure}

\begin{figure}[p]
    \ContinuedFloat
    \centering
    \includegraphics[width=0.95\textwidth]{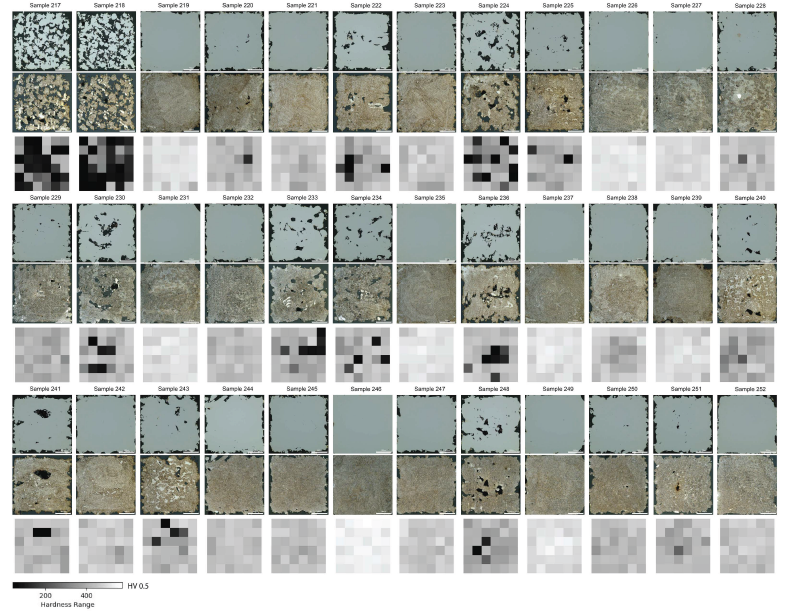}
    \vfill
    \includegraphics[width=0.95\textwidth]{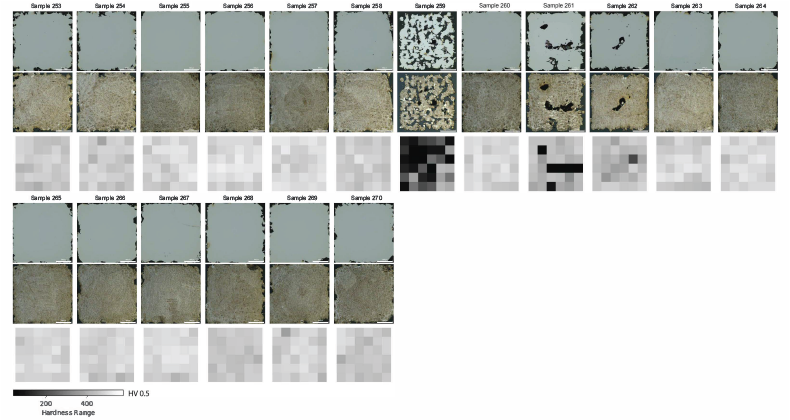}
    \caption{{\textbf{Microstructural images and hardness maps of cuboid samples:} The optical microscope images illustrate the concentration of defects and phase formation, and hardness maps showcase variation of hardness across each cuboid. The figure highlights the large impact of processing conditions on the microstructure, defect content, and property of the samples.}}
    \label{fig: OM_H_all}
\end{figure}

\begin{figure}
    \centering
    \begin{subfigure}[t]{\linewidth}
        \centering
        \includegraphics[width=\linewidth]{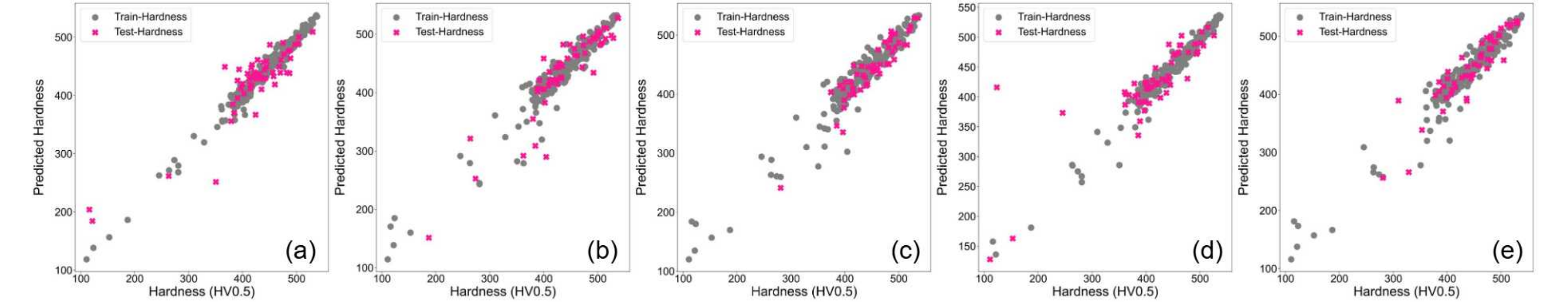}
        \caption{Hardness}
        \label{fig: cv_hardness}
    \end{subfigure}
    \hfill
    \begin{subfigure}[b]{\linewidth}
        \centering
        \includegraphics[width=\linewidth]{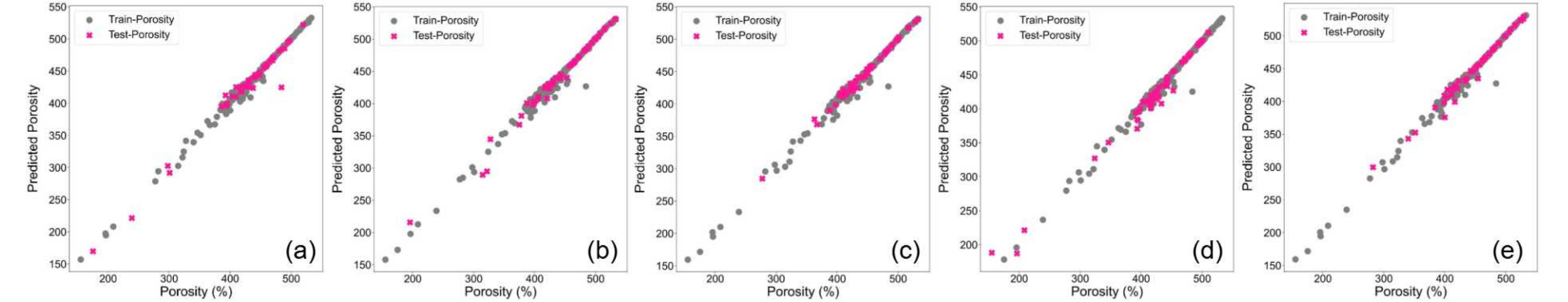}
        \caption{Engineered Porosity}
        \label{fig: cv_porosity}
    \end{subfigure}
    \hfill
    \begin{subfigure}[b]{\linewidth}
        \centering
        \includegraphics[width=\linewidth]{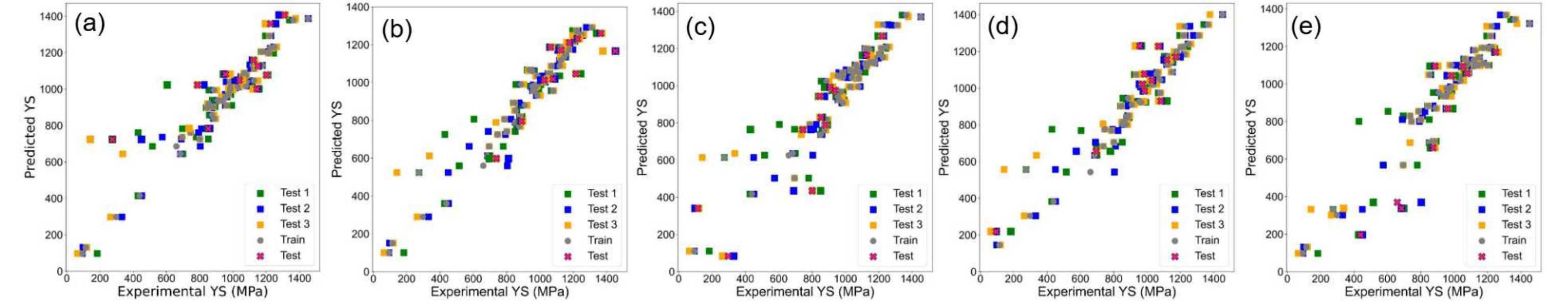}
        \caption{\YS}
        \label{fig: cv_yield}
    \end{subfigure}
    \caption{\textbf{$5$-fold CV of the models trained on hardness, engineered porosity, and \YS}}
    \label{fig: CVs}
\end{figure}


\begin{figure}[p]
    \centering
    \includegraphics[width=\textwidth]{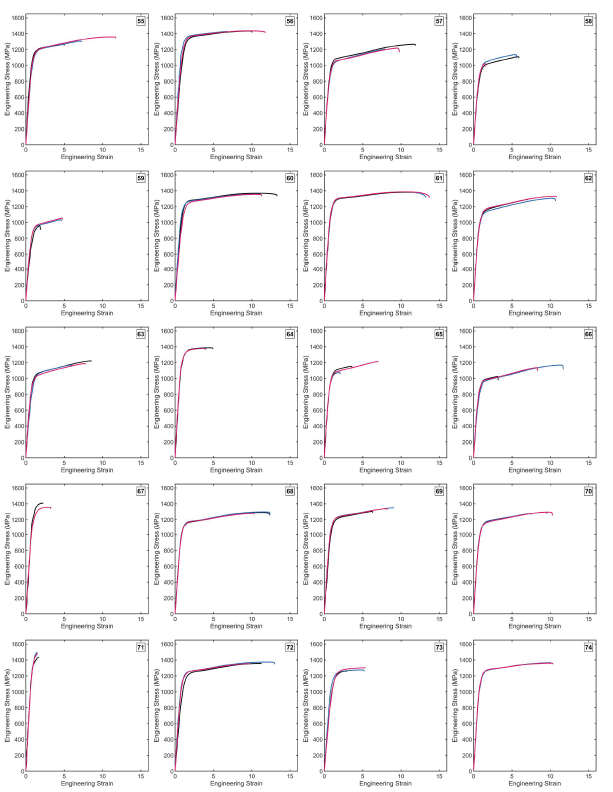}
    \caption[]{}
\end{figure}

\begin{figure}[p]
    \ContinuedFloat
    \centering
    \includegraphics[width=\textwidth]{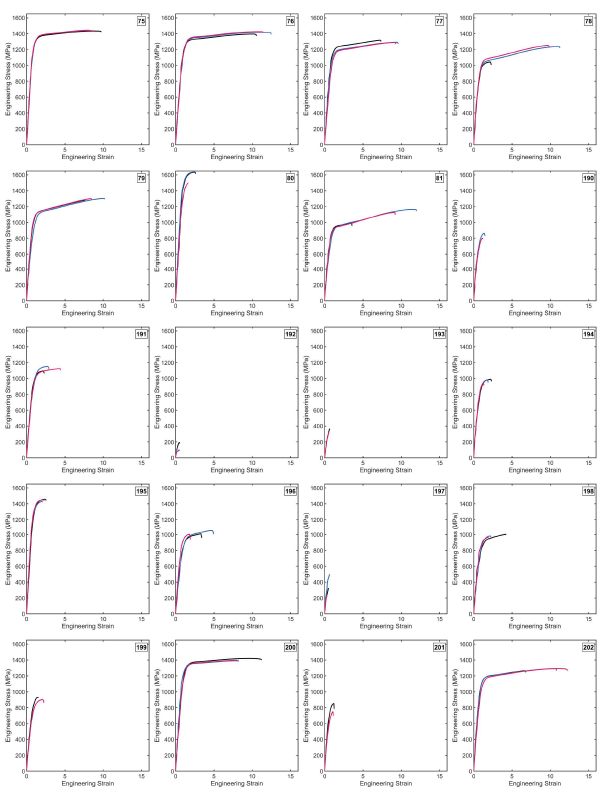}
    \caption[]{}
\end{figure}

\begin{figure}[p]
    \ContinuedFloat
    \centering
    \includegraphics[width=\textwidth]{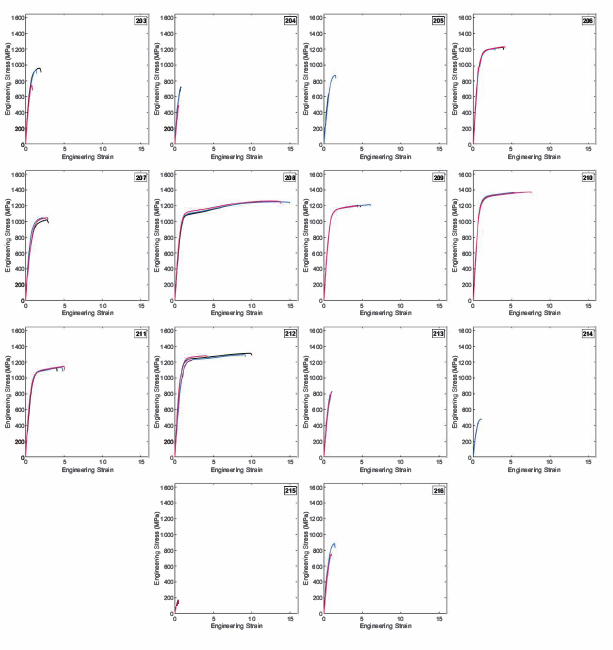}
    \caption{\textbf{Stress-strain curves obtained from the tensile testing of the $54$ processing conditions:} Each plot presents the three curves obtained from the three replicate tensile specimens for each processing condition.}
    \label{fig: stress-strain curves}
\end{figure}



\begin{table}[h!]
    \centering
    \includegraphics[width=\textwidth]{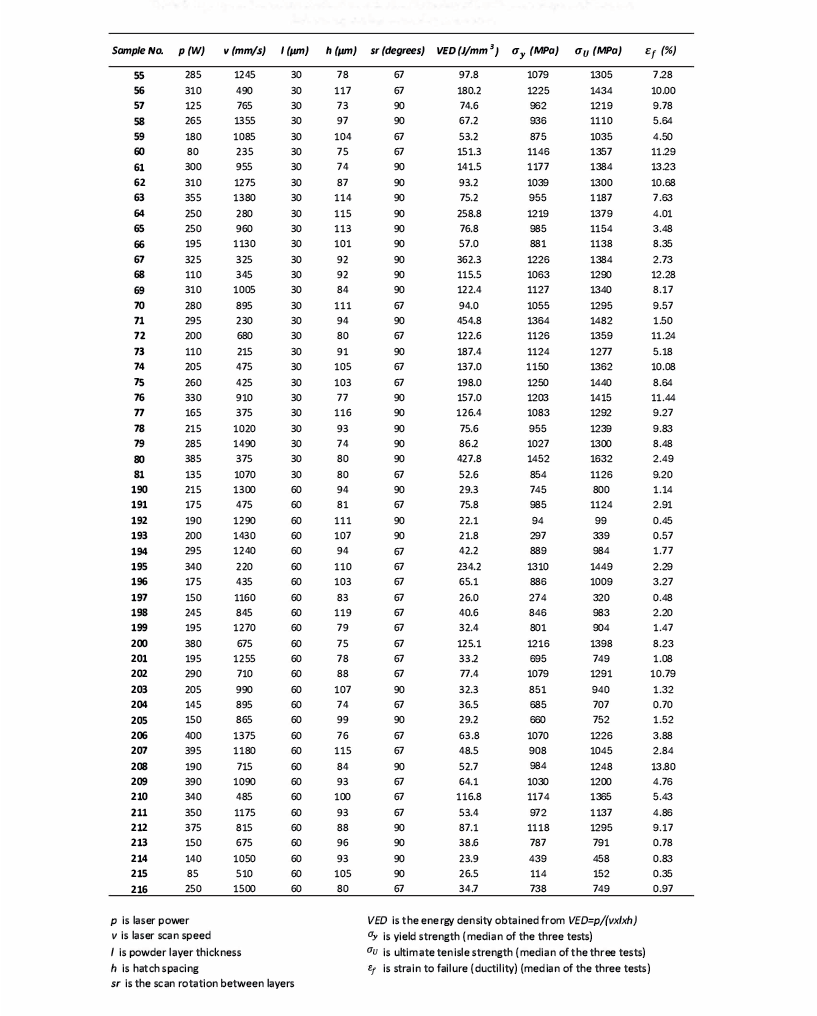}
    \caption{\textbf{\YS, \UTS, and \D~values obtained from the tensile testing of the $54$ processing conditions, and their corresponding processing parameters.}}
    \label{tab: tensile properties}
\end{table}



\begin{figure}
    \centering
    \begin{subfigure}[b]{0.8\linewidth}
        \centering
        \includegraphics[width=\linewidth]{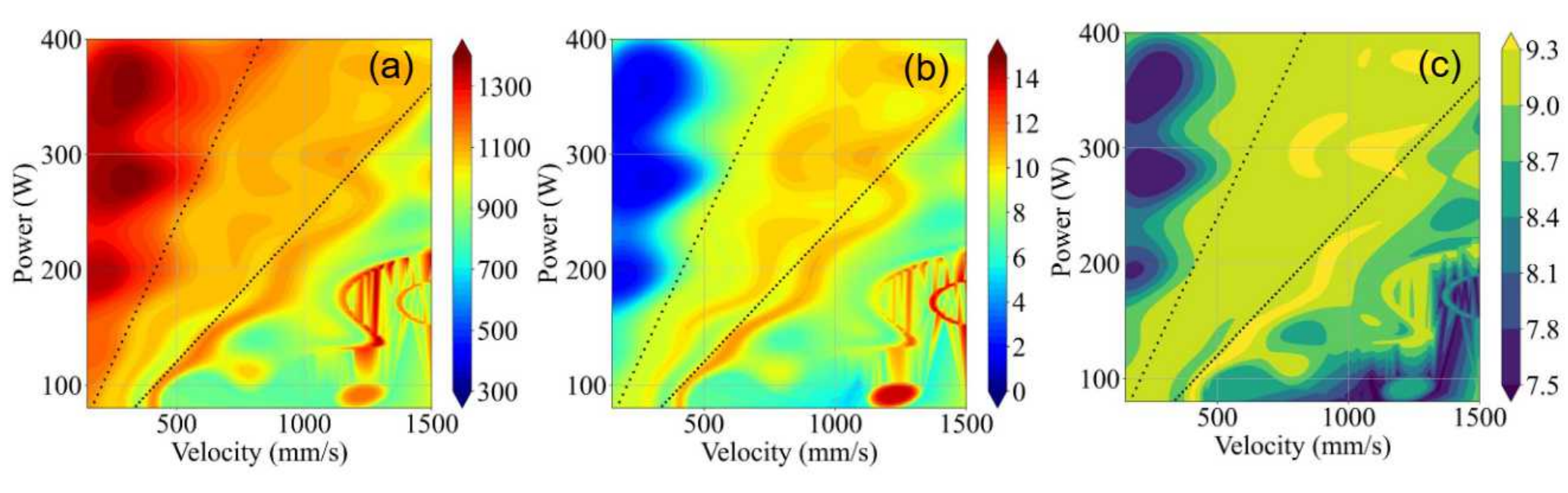}
        \caption{$l=30$ , $h=80$}
        \label{fig: 30_80}
    \end{subfigure}
    \hfill
    \begin{subfigure}[b]{0.8\linewidth}
        \centering
        \includegraphics[width=\linewidth]{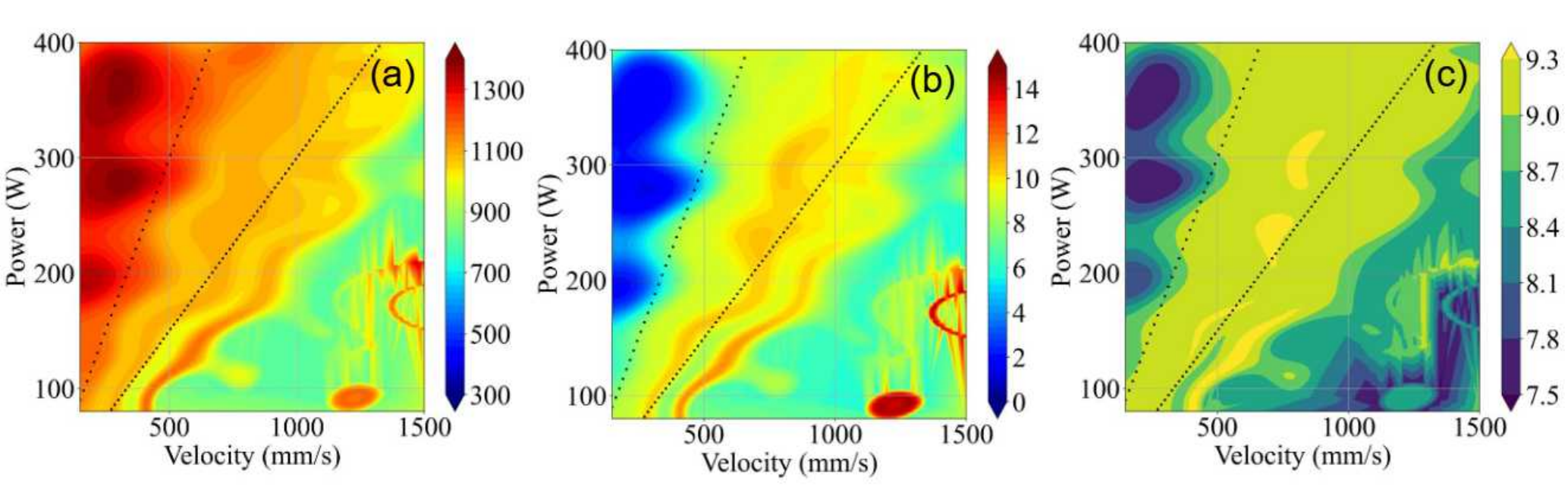}
        \caption{$l=30$ , $h=100$}
        \label{fig: 30_100}
    \end{subfigure}
    \hfill
    \begin{subfigure}[b]{0.8\linewidth}
        \centering
        \includegraphics[width=\linewidth]{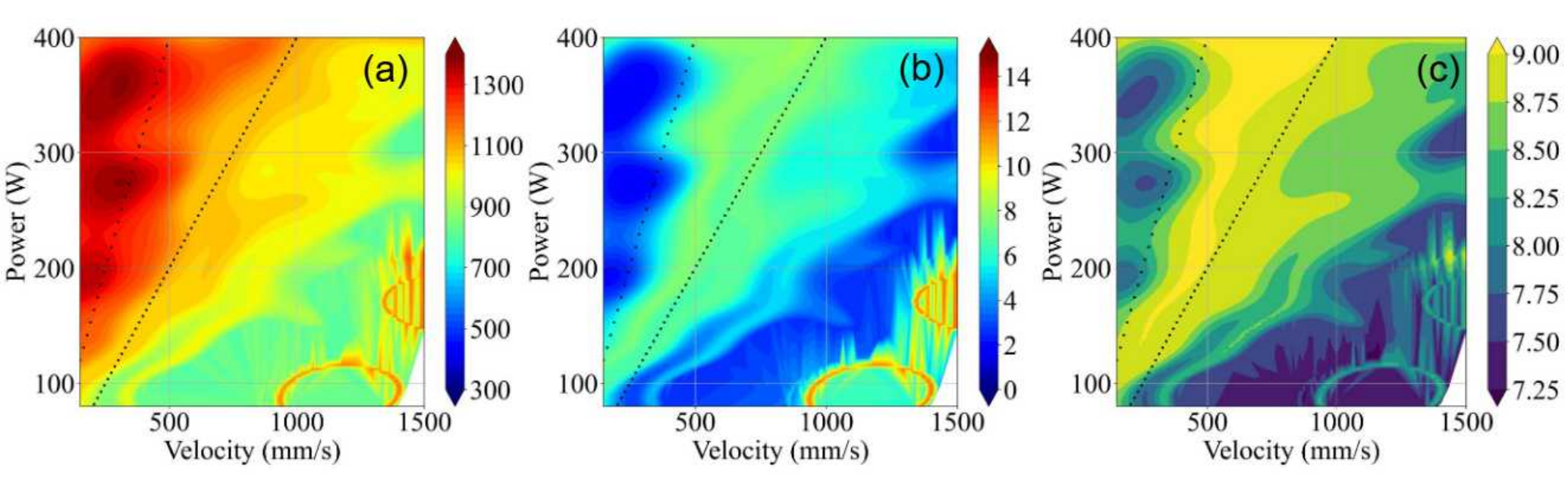}
        \caption{$l=50$ , $h=80$}
        \label{fig: 50_80}
    \end{subfigure}
    \hfill
    \begin{subfigure}[b]{0.8\linewidth}
        \centering
        \includegraphics[width=\linewidth]{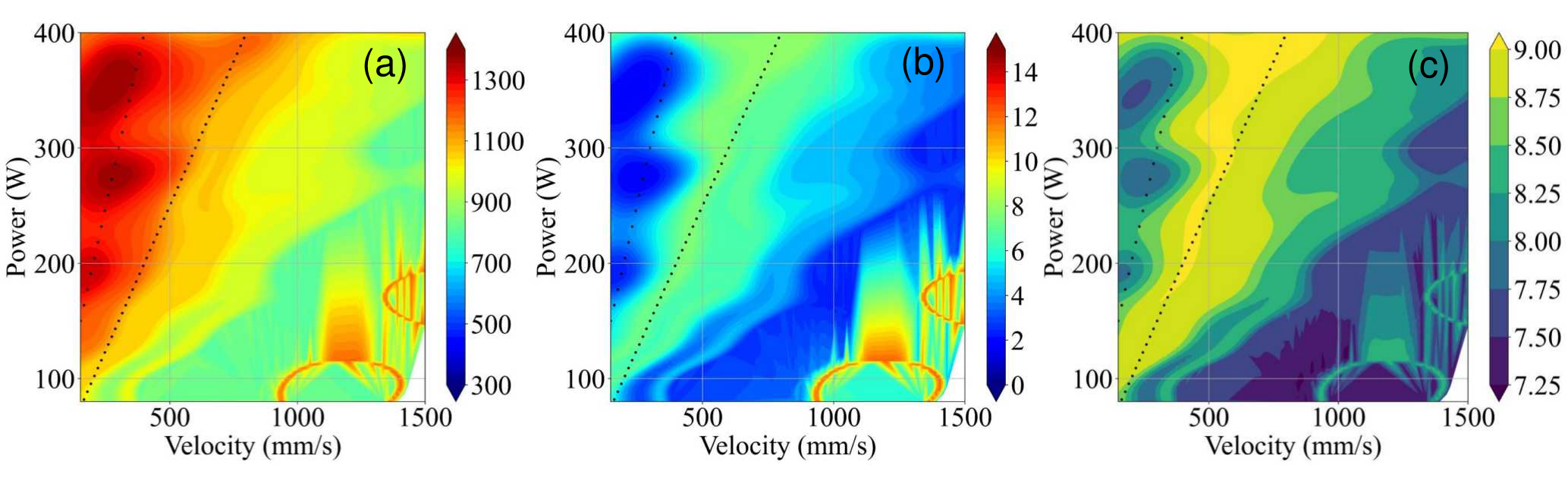}
        \caption{$l=50$ , $h=100$}
        \label{fig: 50_100}
    \end{subfigure}
    \caption{\textbf{Design maps for various combinations of layer thickness and hatch spacing}}
    \label{fig: design_maps_si}
\end{figure}


\end{document}